%% file: main.tex
\pdfoutput=1

\documentclass[11pt]{article}

\usepackage{acl}

\usepackage{times}
\usepackage{latexsym}

\usepackage{graphicx}
\usepackage{subcaption,booktabs}
\usepackage{tabularx}
\usepackage{multirow, amsmath}
\usepackage{placeins}
\usepackage{gensymb} 
\usepackage{amssymb}

\usepackage{amsthm}
\usepackage{color, colortbl}
\usepackage{float}
\usepackage{enumitem}
\usepackage{bbm}
\usepackage{hyperref}

\usepackage{pifont}

\usepackage[T1]{fontenc}

\usepackage[utf8]{inputenc}

\usepackage{microtype}

\usepackage{inconsolata}

\title{Are LLMs Capable of Data-based Statistical and Causal Reasoning? Benchmarking Advanced Quantitative Reasoning with Data}

\author{
	Xiao Liu$^{1}$, Zirui Wu$^{1}$,
        Xueqing Wu$^{2}$, Pan Lu$^{2}$,
	Kai-Wei Chang$^{2}$\and
	Yansong Feng$^{1}$\thanks{\;\;Corresponding author.}\\
        $^1$Wangxuan Institute of Computer Technology, Peking University\\
        $^2$Computer Science Department, University of California, Los Angeles\\
	{\tt \{lxlisa,ziruiwu,fengyansong\}@pku.edu.cn} \\
	{\tt \{xueqing.wu,panlu,kwchang\}@cs.ucla.edu}\\
}

\begin{document}
\maketitle
\begin{abstract} 
Quantitative reasoning is a critical skill to analyze data, yet the assessment of such ability remains limited. To address this gap, we introduce the Quantitative Reasoning with Data (\textsc{QRData}) benchmark, aiming to evaluate Large Language Models' capability in statistical and causal reasoning with real-world data. The benchmark comprises a carefully constructed dataset of 411 questions accompanied by data sheets from textbooks, online learning materials, and academic papers. To compare models' quantitative reasoning abilities on data and text, we enrich the benchmark with an auxiliary set of 290 text-only questions, namely \textsc{QRText}. 
We evaluate natural language reasoning, program-based reasoning, and agent reasoning methods including Chain-of-Thought, Program-of-Thoughts, ReAct, and code interpreter assistants on diverse models. 
The strongest model GPT-4 achieves an accuracy of 58\%, which has much room for improvement. Among open-source models, Deepseek-coder-instruct, a code LLM pretrained on 2T tokens, gets the highest accuracy of 37\%. Analysis reveals that models encounter difficulties in data analysis and causal reasoning, and struggle in using causal knowledge and provided data simultaneously. Code and data are in \url{https://github.com/xxxiaol/QRData}.
\end{abstract}

\input{introduction}
\input{dataset}

\input{methods}
\input{evaluation}
\input{analysis}
\input{related_work}

\input{conclusion}

\input{limitation}

\section*{Acknowledgments}
This work is supported in part by NSFC
(62161160339) and Beijing Science and Technology Program (Z231100007423011). We thank Da Yin and Chen Zhang for their constructive comments. We thank the anonymous reviewers for their helpful discussions and suggestions. For any correspondence, please contact Yansong Feng.

\bibliography{anthology,custom}

\clearpage
\input{appendix}

\end{document}

%% file: introduction.tex
\section{Introduction}
\label{sec-intro}
Data is everywhere in our daily life and work, from win-loss charts in sports competitions to insurance databases of national health. We rely on our \emph{quantitative reasoning skills} to process the data, more specifically, to extract information from the data, draw inferences, and make predictions~\citep{de2021thinking}.

There is evidence that current Large Language Models (LLMs) can perform fundamental data analysis like visualization and summarization~\citep{cheng2023gpt, bai2023qwen}. However, little effort has been made to evaluate the more challenging aspects of quantitative reasoning. 
On the other hand, there are papers evaluating advanced quantitative reasoning abilities of LLMs~\citep{chen-etal-2023-theoremqa, jin2023cladder}, but they have not investigated whether LLMs can apply these skills to real data.

\begin{figure}[t]
    \centering
    \includegraphics[width=0.95\linewidth]{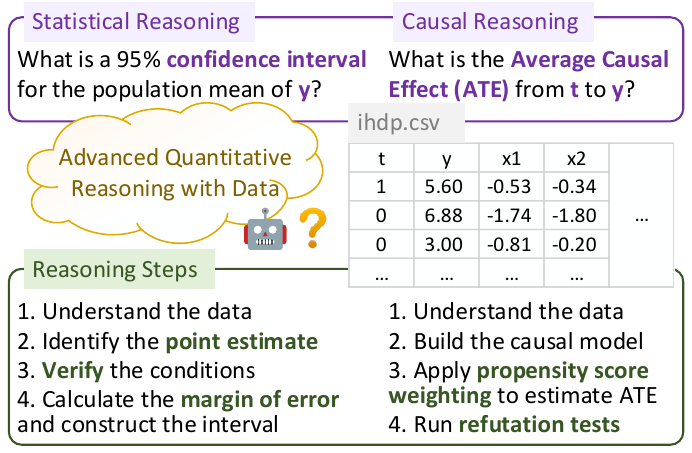}
    \caption{Examples of advanced quantitative reasoning questions and reasoning steps.} 
    \label{fig-intro}
\end{figure}

In this paper, we focus on two main areas of advanced quantitative reasoning: \emph{statistical reasoning} and \emph{causal reasoning}, with examples shown in Figure~\ref{fig-intro}. Given a data sheet from sample surveys, statistical reasoning aims to infer the underlying probability distribution, addressing questions like \emph{what is a 95\% confidence interval for the population mean of $y$}; while causal reasoning aims to understand the causal relationship between variables, addressing questions like \emph{what is the average treatment effect\footnote{Average treatment effect is a measure of causal effect, quantifying the difference in average outcomes between units assigned to the treatment and units assigned to the control.} from $t$ to $y$}.

\begin{figure*}[th]
    \centering
    \includegraphics[width=0.95\textwidth]{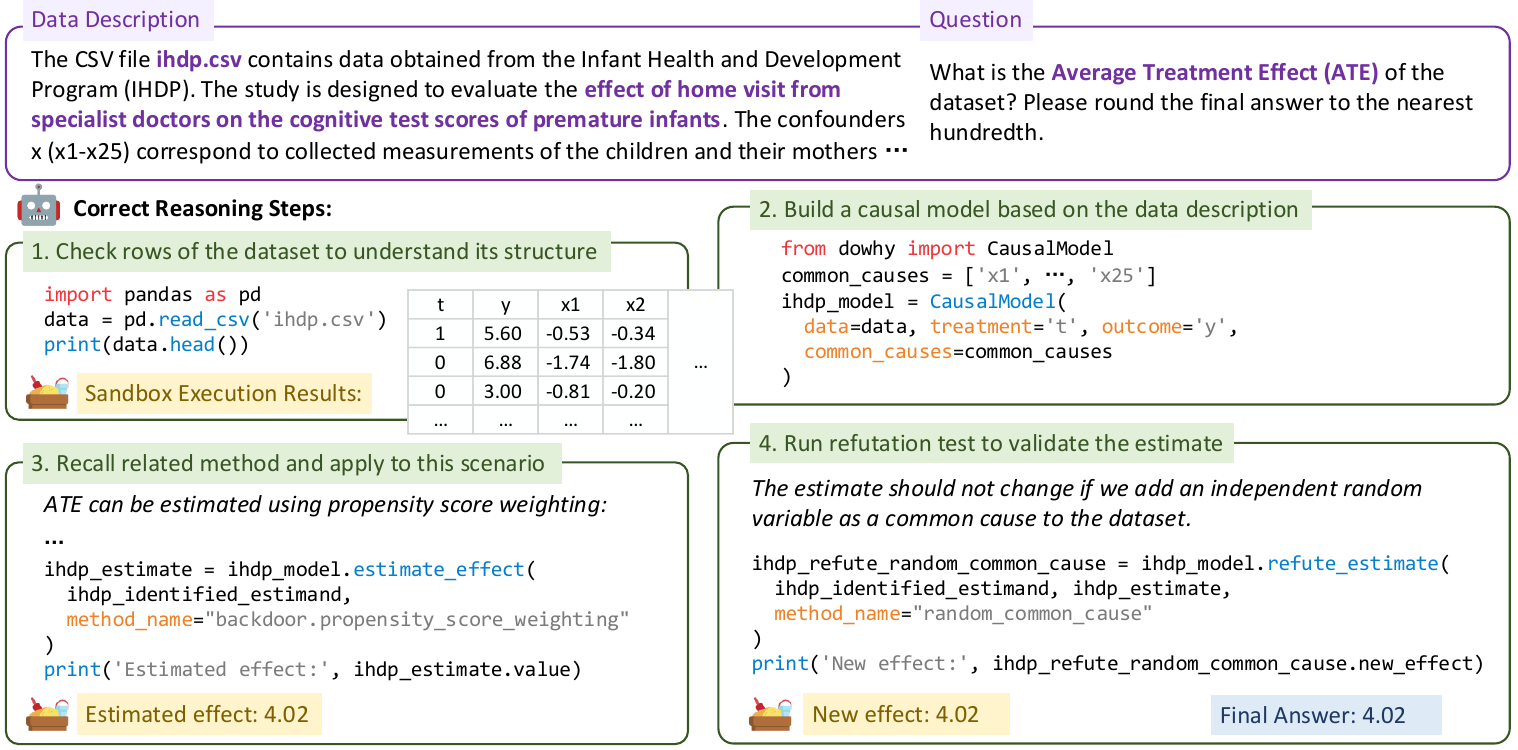}
    \caption{An example question of \textsc{QRData} and correct reasoning steps.} 
    \label{fig-example}
\end{figure*}

We propose the Quantitative Reasoning with Data (\textsc{QRData}) benchmark. It requires models to answer a quantitative reasoning question given data sheets as input. Our benchmark shares a similar form with table question answering~\citep{pasupat-liang-2015-compositional} and data analysis code generation~\citep{lai2023ds} tasks, but has its unique challenges: 
1) Our benchmark requires \emph{advanced quantitative knowledge and skills}. Consider the example in Figure~\ref{fig-example}. To answer the question, models need to know what are common methods of estimating the average treatment effect and how to use them. To the best of our knowledge, few existing benchmarks touch upon concepts in statistical reasoning with data, and none of them involve causal reasoning with data.
2) Our benchmark requires \emph{multi-turn} reasoning. 
After execution of model generated code, models need to read the execution results, and write subsequent code or interpret the results to produce the final answer. In the example, models should first check the data structure before safely indexing the variables in step 2. And after the refutation test in step 4, models need to compare the estimation before and after the refutation, to decide if the estimated effect is reliable, or if another estimation method should be used.

To obtain high-quality quantitative reasoning questions and corresponding data, we construct \textsc{QRData} from open-source textbooks, online learning resources, and academic papers that are accompanied by data. This ensures that all the questions are reasonable and match well with the data. Through careful manual construction, our dataset consists of 411 questions along with 195 data sheets. 
To analyze the difficulty of data analysis, we also collect an auxiliary dataset of Quantitative Reasoning with Text (\textsc{QRText}). \textsc{QRText} contains 290 questions from textbooks and academic papers, and the questions can be solved based on text descriptions only.

We evaluate a wide range of models and reasoning approaches on \textsc{QRData}. To find what is a good solution to the task, we implement natural language reasoning with Chain-of-Thought~\citep{wei2022chain}, program-based reasoning with Program-of-Thoughts~\citep{chen2022program}, agent reasoning with ReAct~\citep{yao2022react} and the code interpreter assistants of GPT models. We also compare general-purpose LLMs with LLMs finetuned for specific abilities like math, code, and agent reasoning, to find out whether these abilities help in tackling our benchmark. Experimental results show that the best closed-source LLM GPT-4~\citep{openai2023gpt4} achieves an accuracy of 58\% with the code interpreter assistant, and the best open-source LLM Deepseek-coder-instruct~\citep{guo2024deepseek} achieves 37\% with Program-of-Thoughts.

As there is much room for improvement in model performance, we further dive into what are the primary difficulties. We find that data analysis is challenging for LLMs, especially for smaller open-source LLMs. And all models including GPT-4 and Gemini~\citep{team2023gemini} struggle with causal reasoning. They acquire causal knowledge from the pretraining corpus, but can hardly integrate the knowledge with the provided data in reasoning.

Our main contributions are as follows: 
1) We propose \textsc{QRData}, the first benchmark for advanced quantitative reasoning with data, to assess models' abilities of data-based statistical and causal reasoning.
2) We evaluate various models and reasoning approaches on \textsc{QRData}, and provide insights into effective approaches to tackle the benchmark.
3) We analyze the difficulties of current models in data analysis and causal reasoning. 

%% file: dataset.tex
\section{Benchmark Construction}
In this section, we introduce how the \textsc{QRData} benchmark is constructed.
To ensure the quality of our benchmark, we first gather teaching and research resources for quantitative reasoning, and then annotate questions based on the materials. To compare the performance of quantitative reasoning on data and text, we also introduce an auxiliary dataset \textsc{QRText} for analysis.

\vspace{2mm}
\noindent\textbf{Resource Selection.} We look for quantitative reasoning resources with the following criteria: 
1) The resource is related to quantitative reasoning, especially statistical reasoning and causal reasoning.
2) It contains multiple quantitative reasoning questions with corresponding data and answers.
3) It is open-source and allows adaptation.

Eight resources are selected based on these criteria, including two books: OpenIntro Statistics~\citep{diez2012openintro} and Quantitative Social Science~\citep{imai2018quantitative}; one online learning resource: Causal Inference for the Brave and True\footnote{https://matheusfacure.github.io/python-causality-handbook/landing-page.html}; and five datasets from academic papers: Flow cytometry~\citep{sachs2005causal}, IHDP~\citep{hill2011bayesian}, Jobs~\citep{shalit2017estimating}, Neuropathic pain diagnosis~\citep{tu2019neuropathic}, and Arctic sea ice~\citep{huang2021benchmarking}. 

\vspace{2mm} 
\noindent\textbf{Question Collection.}
We collect multiple-choice questions and numerical questions from these resources, and ensure that the gold answer is unique.
For textbooks and online learning resources, we extract the list of data sheets from their websites. If a data sheet is used in exercises, we use the exercise question and its solution as the question and answer; if it is used in the main content, we annotate questions and answers based on the content.

For existing causal discovery datasets (Flow cytometry, Neuropathic pain diagnosis, and Arctic sea ice), we collect multiple-choice questions like \emph{Which cause-and-effect relationship is more likely? A. headache causes groin pain B. groin pain causes headache C. No causal relationship exists} following the style of~\citet{kiciman2023causal}. 
For causal estimation datasets (IHDP and Jobs), we collect numerical questions of estimating \emph{average treatment effect (ATE)}, \emph{average treatment effect on the treated (ATT)}, and \emph{average treatment effect on the control (ATC)} like the example in Figure~\ref{fig-example}. More details are in Appendix~\ref{sec-appendix-qrdata}.

We also collect data descriptions like the purposes and formats of the data sheets from the resources, and provide them along with questions to models.

\input{tables/dataset_statistics}
\begin{figure}[t]
    \centering
    \includegraphics[width=0.9\linewidth]{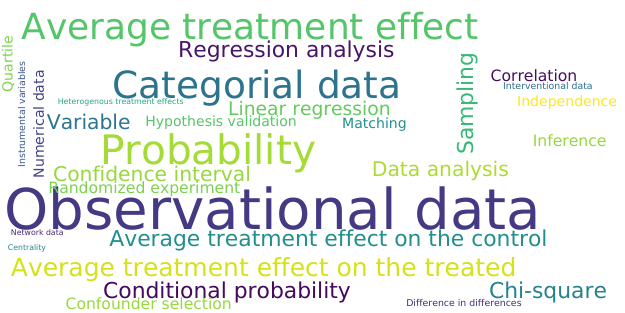}
    \caption{Key concepts in \textsc{QRData}.} 
    \label{fig-keyword}
\end{figure}

\vspace{2mm} 
\noindent\textbf{Statistics.} We provide the dataset statistics in Table~\ref{table-dataset-statistics}, and the number of questions collected from each source in Appendix Table~\ref{table-dataset-source}. The questions we collect cover a wide range of concepts, and we illustrate some of them in Figure~\ref{fig-keyword}. We also summarize the topics and question examples in Appendix Table~\ref{table-topic-example}. Most questions in \textsc{QRData} are based on a single data sheet, while 4\% of questions are based on multiple data sheets.

To check whether our dataset has already been memorized by models in training, we apply the pretraining data detection method Min-K\% Prob~\citep{shi2023detecting} on GPT models and Llama-2, and confirm that our dataset is unlikely to appear in their pretraining corpora.

\vspace{2mm} 
\noindent\textbf{\textsc{QRText}.} To separate the challenge of quantitative reasoning from data analysis and analyze whether models master the quantitative reasoning skills, we create an auxiliary benchmark for comparison called Quantitative Reasoning with Text. 

Questions in \textsc{QRText} can be answered without data, with an example shown in Appendix Figure~\ref{fig-qrtext}. We collect questions from OpenIntro Statistics and two text-based causal reasoning datasets \textsc{CLadder}~\citep{jin2023cladder} and \textsc{Corr2Cause}~\citep{jin2023can}. \textsc{QRText} contains 290 questions, 100 for statistical reasoning and 190 for causal reasoning. For ease of model comparison, the ratio of statistical/causal questions of \textsc{QRText} is similar to \textsc{QRData}. Details of \textsc{QRText} question collection are in Appendix~\ref{sec-appendix-qrtext}.

\vspace{2mm}
\noindent\textbf{Evaluation Metric.} We report the accuracy by comparing the final answer generated by models with the gold answer. As models may encounter measurement errors in numerical questions (e.g., different causal effect estimation methods will yield minor differences in results), we consider any result within $\pm 3\%$ of the gold answer to be correct.

%% file: tables/dataset_statistics.tex
\begin{table}[t]
\centering
\small
    \begin{tabular}{lc}
        \toprule
        & \textbf{Count} \\
        \midrule
        \rowcolor[gray]{0.95} Data Files & 195 \\
        \quad Avg. Rows per File & 15219 \\
        \quad Avg. Columns per File & 46 \\
        \quad Avg. Words of Data Description & 73 \\
        \rowcolor[gray]{0.95} Questions & 411 \\
        \quad Multiple Choice / Numerical & 248 / 163 \\
        \quad Statistical / Causal & 142 / 269 \\
        \quad Avg. Words of Question & 32 \\
        \bottomrule
    \end{tabular}
    \caption{Statistics of \textsc{QRData}.}
    \label{table-dataset-statistics}
\end{table}

%% file: methods.tex
\section{Methods}
We develop several zero-shot reasoning methods as our baselines. By comparing these methods, we aim to find a good solution to the task of quantitative reasoning with data. 

\vspace{2mm}
\noindent\textbf{Table Question Answering (Table QA).} The task of table QA asks models to answer a question based on tables. To evaluate if our benchmark can be solved by existing table QA methods, we experiment with a competitive Table QA method~\citep{zhang2023tablellama}. It constructs a table instruction tuning dataset \texttt{TableInstruct}, and finetunes models on this dataset.

\vspace{2mm}
\noindent\textbf{Chain-of-Thought (CoT).} CoT prompting enables models to conduct complex natural language reasoning through intermediate reasoning steps. We feed the content of data sheets to models and prompt models to \texttt{respond to a quantitative question using the provided data ... Let's think step by step}. If the data sheet is too large, we truncate it to fit the length limit. To obtain the final answer from the model output, we employ ChatGPT for answer extraction following~\citet{chen-etal-2023-theoremqa}.\footnote{To assess the reliability of answer extraction with ChatGPT, we randomly sample 50 CoT outputs from all models, and manually verify the extracted answers. ChatGPT achieves an accuracy of 98\%, demonstrating its proficiency in answer extraction.}

\vspace{2mm}
\noindent\textbf{Program-of-Thoughts (PoT).} PoT prompting asks models to generate a Python code program, and uses an external computer to execute the code. The output of the code is regarded as the answer. In contrast to TableLlama and CoT prompting, PoT processes the data sheet with code and gets rid of the length limit of data. It also enables models to call data analysis packages for computation. 

\vspace{2mm}
\noindent\textbf{ReAct-style Prompting.} ReAct is a prevalent prompting paradigm for agent reasoning. It combines reasoning and acting in language models for task solving. To address our task, we restrict the action space to Python code execution, and ask models to generate \emph{thoughts} and \emph{actions}.
Different from aforementioned methods that only allow one-turn generation, ReAct supports multi-turn reasoning. When a piece of code is generated, we execute it and feed the output back to models as an \emph{observation}. Based on the observation, models will either generate new thoughts and actions, or decide to predict the final answer.

\vspace{2mm}
\noindent\textbf{Code Interpreter Assistants.} GPT models can be used as agents (called \emph{assistants}), and a code interpreter tool is built-in, which could execute Python code in a sandbox to interpret the data. Similar to ReAct-style prompting, code interpreter assistants allow multi-turn code generation and execution, and end with a text-formed answer.

Details of prompt design and prompt examples are in Appendix~\ref{sec-appendix-prompt}.

%% file: evaluation.tex
\section{Experiments}
\input{tables/main_results}
\subsection{Models}
We experiment with the CoT, PoT, and ReAct-style prompting methods on the following models:

\vspace{2mm}
\noindent\textbf{General-Purpose LLMs.} We select Llama-2-chat, Gemini-Pro, and GPT-4 as representatives of general-purpose LLMs. They exhibit strong abilities to follow instructions~\citep{touvron2023llama, team2023gemini, openai2023gpt4}.

\vspace{2mm}
\noindent\textbf{Code LLMs.} CodeLlama-instruct~\citep{roziere2023code} is initialized on Llama-2, finetuned on a code-heavy dataset, and instruction tuned on a code-related instruction dataset. Deepseek-coder-instruct~\citep{guo2024deepseek} is pretrained on a project-level code corpus from scratch and instruction tuned to follow instructions. Both of them perform well among open-source code models on several code benchmarks.

\vspace{2mm}
\noindent\textbf{LLM for Mathematical Reasoning.} WizardMath~\citep{luo2023wizardmath} is finetuned on Llama-2, enhancing the mathematical reasoning abilities with reinforcement learning. It outperforms many larger models on mathematical reasoning benchmarks.

\vspace{2mm}
\noindent\textbf{LLM for Agent Reasoning.} AgentLM~\citep{zeng2023agenttuning} is instruction-tuned based on Llama-2 to enhance the agent capabilities. It uses a combination of instructions from general domains and instructions of interaction trajectories.

For a fair comparison, we use the size of 7B for all the open-source models.\footnote{The size of Deepseek-coder-instruct is 6.7B.}
We use the two latest GPT models GPT-3.5 Turbo and GPT-4\footnote{The specific versions are \texttt{gpt-3.5-turbo-1106} and \texttt{gpt-4-1106-preview}.} for code interpreter assistants.
More implementation details are in Appendix~\ref{sec-appendix-implementation}.

We select TableLlama~\citep{zhang2023tablellama} as a representative table QA model. It is finetuned on Llama-2-7B~\citep{touvron2023llama} with \texttt{TableInstruct}, using LongLoRA~\citep{chen2023longlora} to extend the context length. It exhibits strong generalization ability across table QA datasets.

As a reference for assessment, we also report the random baseline and the human performance. The random baseline is calculated by randomly choosing from the multiple choices, and all the numerical questions are counted as incorrect. 

The human performance is evaluated by asking three graduate students who have taken quantitative reasoning courses to answer a random sample of 50 questions from \textsc{QRData}. Each question is answered by at least two students, and the average accuracy is reported. The students are allowed to use a Python executor, and search for concepts rather than specific questions on the Internet. \footnote{The majority of human errors are in causal discovery questions (71\%), because state-of-the-art causal discovery algorithms cannot fully resolve this task~\citep{vowels2022d}.}

\subsection{Main Results}
To investigate which reasoning methods and models better solve our task, we summarize the performance of different models paired with different reasoning methods in Table~\ref{table-main-results}. GPT-4 with the code interpreter assistant achieves the best performance, and Deepseek-coder-instruct with PoT prompting is the best among open-source models. The best model is 18\% worse than human, showing that \textsc{QRData} is challenging for LLMs.

\vspace{2mm}
\noindent\textbf{Natural Language Reasoning.} CoT works best for several open-source models like Llama-2-chat and WizardMath, but the results are close to or even worse than random, as data-based quantitative reasoning requires heavy calculations that are beyond natural language reasoning. The table QA model TableLlama also underperforms across both statistical and causal tasks.

Models can hardly answer numerical questions with CoT, with a maximum accuracy of less than 10\%. A limitation of CoT and other natural language reasoning methods is that in many cases, we cannot feed the whole data sheets into the input, hindering models from performing accurate calculations. 
We further compare CoT and PoT prompting on a subset of \textsc{QRData} where the full data sheets can be fed into the base model, and find that CoT does not exceed PoT on this subset. The accuracy of CoT prompting on numerical questions is below 20\% across all models, suggesting these models struggle with performing precise calculations on data sheets using CoT.

\vspace{2mm}
\noindent\textbf{Program-based Reasoning.} PoT best suits Deepseek-coder-instruct, but to our surprise, another code LLM CodeLlama-instruct does not perform well with PoT. The gap in coding abilities can also be seen from the execution success rate, which is 89\% for Deepseek-coder-instruct and 73\% for CodeLlama-instruct. 

Although the overall performance of several non-code LLMs drops with PoT compared to CoT, the accuracy of numerical questions improves with the help of code execution. This exhibits the potential for performance improvement through the hybrid of CoT and PoT rationales~\citep{yue2023mammoth}.

\vspace{2mm}
\noindent\textbf{Agent Reasoning.} Closed-source LLMs like GPT-4 and Gemini-Pro perform best with agent reasoning. GPT-4 performance with ReAct-style prompting is close to with the code interpreter assistant, exhibiting that our ReAct-style prompts successfully activate the model abilities.

Smaller open-source models can hardly follow the complex ReAct-style instructions, which limits their performance. Although AgentLM is finetuned for agent reasoning, it cannot generalize well to the quantitative reasoning task. On the other hand, open-source models except code LLMs perform better with ReAct-style prompting than PoT, because multi-turn code execution gives them chances to fix the generated code, raising the average execution success rate from 25.5\% to 35.6\%. We also manually review 50 randomly sampled ReAct outputs of these models, and find that in 10\% of the cases, the initially generated code is non-executable, and is fixed in the following steps. This suggests that the iterative, multi-turn nature of agent reasoning can contribute to fixing the generated code. 

\vspace{2mm}
\noindent\textbf{LLMs with Enhanced Abilities.} 
We compare three models finetuned on Llama-2 for enhanced abilities of code, math, and agent reasoning with the general-purpose LLM Llama-2-chat. AgentLM and WizardMath only gain marginal improvement, and CodeLlama-instruct even underperforms Llama-2-chat. This may be due to the limited capability of the base model. 

On the other hand, Deepseek-coder-instruct, which masters in code generation, outperforms all other models with similar sizes by more than 10\%. We further evaluate Deepseek-coder-instruct-33B to see if it could approach the performance of closed-source LLMs, but it gets an overall accuracy of 31.9\% with PoT and 27.0\% with ReAct-style prompting, even lower than the 6.7B model.
Although coding ability is important in solving quantitative questions with data, the performance of code LLMs is still limited compared to larger general-purpose LLMs like GPT-4. 

\vspace{2mm}
\noindent\textbf{Statistical Reasoning vs. Causal Reasoning.}
Statistical reasoning focuses on identifying patterns in data, whereas causal reasoning goes a step further to find causal relationships beyond correlation.
For all models evaluated, the gap between model performance and the random baseline is larger for statistical questions than causal questions, showing that models obtain better statistical reasoning abilities than causal reasoning abilities. Even the best-performed model GPT-4 only answers 51\% of causal questions correctly, demonstrating the large room for improvement in causal reasoning.

%% file: tables/main_results.tex
\begin{table*}[t]
\centering
\small
\scalebox{0.97}{
\begin{tabular}{lcccc}
    \toprule
    \textbf{Model} & \textbf{Size} & \textbf{Overall Acc.} & \textbf{Multiple Choice / Numerical} & \textbf{Statistical / Causal} \\
    \midrule
    Random &  & 23.0 & 38.1 / 0.0 & 15.0 / 27.2 \\
    \rowcolor[gray]{0.95} \textit{Table Question Answering} &  &  &  &  \\
    \quad TableLlama~\citep{zhang2023tablellama} & 7B & 9.5 & 14.9 / 1.2 & 3.5 / 12.6 \\
    \rowcolor[gray]{0.95} \textit{Chain of Thought Prompting} &  &  &  &  \\
    \quad Llama-2-chat~\citep{touvron2023llama} & 7B & \underline{22.1} & \underline{35.1} / \underline{2.5} & \underline{20.4} / \underline{23.0} \\
    \quad AgentLM~\citep{zeng2023agenttuning} & 7B & \underline{24.3} & \underline{40.0} / 0.0 & \underline{19.0} / \underline{27.1} \\
    \quad WizardMath~\citep{luo2023wizardmath} & 7B & \underline{25.1} & \underline{40.0} / 2.5 & 18.3 / \underline{28.6} \\
    \quad CodeLlama-instruct~\citep{roziere2023code} & 7B & \underline{19.2} & \underline{31.0} / 1.2 & 14.1 / \underline{21.9} \\
    \quad Deepseek-coder-instruct~\citep{guo2024deepseek} & 6.7B & 18.7 & 30.2 / 1.2 & 15.5 / 20.4 \\
    \quad Gemini-Pro~\citep{team2023gemini} & - & 31.9 & 46.8 / 9.2 & 25.4 / 35.3 \\
    \quad GPT-4~\citep{openai2023gpt4} & - & 36.5 & 57.3 / 4.9 & 24.6 / 42.8 \\
    \rowcolor[gray]{0.95} \textit{Program of Thoughts Prompting} &  &  &  &  \\
    \quad Llama-2-chat & 7B & 3.7 & 6.1 / 0.0 & 7.8 / 1.5 \\
    \quad AgentLM & 7B & 3.2 & 4.4 / \underline{1.2} & 8.5 / 0.4 \\
    \quad WizardMath & 7B & 16.8 & 16.5 / \underline{17.2} & \underline{31.7} / 8.9 \\
    \quad CodeLlama-instruct & 7B & 17.3 & 24.2 / \underline{6.7} & \underline{19.7} / 16.0 \\
    \quad Deepseek-coder-instruct & 6.7B & \underline{37.0} & \underline{40.7} / \underline{31.3} & \underline{45.8} / \underline{32.3} \\
    \quad Gemini-Pro & - & 26.8  & 27.8 / 25.2 & 39.4 / 20.1 \\
    \quad GPT-4 & - & 49.1 & 46.0 / 54.0 & 72.5 / 36.8 \\
    \rowcolor[gray]{0.95} \textit{ReAct-style Prompting} &  &  &  &  \\
    \quad Llama-2-chat & 7B & 11.4 & 18.5 / 0.6 & 4.2 / 15.2 \\
    \quad AgentLM & 7B & 10.7 & 16.9 / \underline{1.2} & 5.6 / 13.4 \\
    \quad WizardMath & 7B & 18.7 & 21.0 / 15.3 & 19.7 / 18.2 \\
    \quad CodeLlama-instruct & 7B & 15.8 & 22.2 / 6.1 & 16.2 / 15.6 \\
    \quad Deepseek-coder-instruct & 6.7B & 27.3 & 29.4 / 23.9 & 38.7 / 21.2 \\
    \quad Gemini-Pro & - & \underline{38.9} & \underline{47.6} / \underline{25.8} & \underline{41.5} / \underline{37.5} \\
    \quad GPT-4 & - & 57.4 & \underline{\textbf{59.2}} / 54.6 & 69.0 / \underline{\textbf{51.3}} \\
    \rowcolor[gray]{0.95} \textit{Code Interpreter Assistants} &  &  &  &  \\
    \quad GPT-3.5 Turbo & - & 41.4 & 42.3 / 39.9 & 60.6 / 31.2 \\
    \quad GPT-4 & - & \underline{\textbf{57.9}} & 56.9 / \underline{\textbf{59.5}} & \underline{\textbf{78.9}} / 46.8 \\
    \rowcolor[gray]{0.95} Human (subset) &  & 76.0 & 71.9 / 83.4 & 88.8 / 68.8 \\
    \bottomrule
\end{tabular}
}
\caption{Performance of models on \textsc{QRData}. Numbers are accuracies in percentages (\%). The best results are in bold. For models evaluated with multiple reasoning methods, the model-level best results are underlined.}
\label{table-main-results}
\end{table*}

%% file: analysis.tex
\section{Analysis}
\input{tables/results_no_data}
\input{tables/results_multiple_choice_without_data}
\input{tables/case}
We go deeper into the primary difficulties models face in addressing the task of quantitative reasoning with data. This may provide insights into how to design methods to better solve our task. 
We also provide an error analysis of GPT-4 in Appendix~\ref{sec-appendix-error}.
\subsection{Difficulty of Data Analysis}
We evaluate models\footnote{We do not experiment with TableLlama as the questions do not involve tables.} on our auxiliary benchmark \textsc{QRText}, and compare the performance on \textsc{QRText} and \textsc{QRData} to quantify the difficulty of data analysis.
As shown in Table~\ref{table-results-no-data}, GPT-4 achieves the best accuracy of 59.7\% on \textsc{QRText}, and Deepseek-coder-instruct gets the best accuracy of 39.0\% among open-source LLMs.

The single-turn reasoning methods, CoT and PoT prompting, work best for all models except GPT-4 on \textsc{QRText}. For GPT-4, the performance gap between the code interpreter assistant and PoT is merely 0.7\%, compared to 8.8\% on \textsc{QRData}. These demonstrate that single-turn reasoning is almost adequate for quantitative reasoning with text, while multi-turn reasoning is more necessary for quantitative reasoning with data.

All models perform better on \textsc{QRText} than \textsc{QRData} from 1.8\% to 11.5\%, and the gap is larger for smaller models. If we control the knowledge and skills required by restricting questions to the same source (OpenIntro Statistics), models perform 6\% better on average. These provide evidence that most models have difficulty in data analysis.

\subsection{Difficulty of Causal Reasoning}
We observe the performance gap between statistical reasoning and causal reasoning on \textsc{QRData}, and the gap remains on \textsc{QRText}. GPT-4 achieves 89\% accuracy in statistical questions on \textsc{QRText}, but only about half the accuracy in causal questions. This exhibits the unique difficulty of causal reasoning regardless of data analysis.
 
Table~\ref{table-case} provides a case where GPT-4 with the code interpreter assistant mis-predicts a causal question from \textsc{QRData}. When asked to predict the causal relation between two variables \emph{L L1 radiculopathy} and \emph{R L5 radiculopathy}, GPT-4 proposes a wrong plan for calculating the conditional probabilities in Step 2. As correlation does not imply causation, although GPT-4 successfully executes the plan, it makes a wrong prediction. 

\vspace{2mm}
\noindent\textbf{Easier Said than Done.}
The unsatisfactory performance on causal questions appears to differ from the findings of some previous papers~\citep{willig2022can, kiciman2023causal}, which show that LLMs are capable of causal reasoning in tasks like causal discovery \emph{without data}. To figure out the reason for the inconsistency, we conduct an experiment to evaluate if the provided data interfered with the models. Specifically, we evaluate the best three models on multiple-choice causal questions of \textsc{QRData} \emph{without giving them the data}\footnote{Note that this setting is different with \textsc{QRText}. Without data access, questions in \textsc{QRData} are not self-contained. Models need to predict the answers with their commonsense.}, and compare their performance with the original setting. 
We use the best reasoning method for each model: ReAct-style prompting for Gemini-Pro, and code interpreter assistants for GPT-3.5 Turbo and GPT-4. 

As shown in Table~\ref{table-multiple-choice-wo-data}, two out of three models perform better without data access, revealing that it is easier for these LLMs to speculate causal relations than to conduct causal reasoning with data.
The strong causal reasoning abilities exhibited by models in previous works come from their internal knowledge, but models can hardly use their internal knowledge to analyze the data provided.

In the case of Table~\ref{table-case}, in the absence of data access, GPT-4 correctly predicts that there is no causal relationship between the two variables. However, when the data is provided, it is misled to calculate the conditional probabilities. 

%% file: tables/results_no_data.tex
\begin{table}[t]
\centering
\small
\begin{tabular}{lcc}
    \toprule
    \textbf{Model} & \textbf{Overall Acc.} & \textbf{Statistical / Causal} \\
    \midrule
    Random & 23.7 & 20.8 / 25.3 \\
    \rowcolor[gray]{0.95} \multicolumn{3}{l}{\textit{Chain of Thought Prompting}}\\
    \quad Llama-2-chat & \underline{28.6} & \underline{33.0} / 26.3 \\
    \quad AgentLM & \underline{26.6} & \underline{28.0} / \underline{25.8} \\
    \quad WizardMath & \underline{36.6} & \underline{46.0} / 31.6 \\
    \quad CodeLlama-i & 26.2 & \underline{26.0} / 26.3 \\
    \quad Deepseek-coder-i & 23.1 & 18.0 / 25.8 \\
    \quad Gemini-Pro & 37.9 & 51.0 / 31.1 \\
    \quad GPT-4 & 52.1 & 76.0 / 39.5 \\
    \rowcolor[gray]{0.95} \multicolumn{3}{l}{\textit{Program of Thoughts Prompting}}\\
    \quad Llama-2-chat & 21.7 & 12.0 / \underline{26.8} \\
    \quad AgentLM & 22.4 & 16.0 / \underline{25.8} \\
    \quad WizardMath & 34.1 & 38.0 / \underline{32.1} \\
    \quad CodeLlama-i & \underline{27.2} & \underline{26.0} / \underline{27.9} \\
    \quad Deepseek-coder-i & \underline{39.0} & \underline{47.0} / \underline{34.7} \\
    \quad Gemini-Pro & \underline{43.1} & \underline{56.0} / \underline{36.3} \\
    \quad GPT-4 & 59.0 & 86.0 / \underline{\textbf{44.7}} \\
    \rowcolor[gray]{0.95} \multicolumn{3}{l}{\textit{ReAct-style Prompting}}\\
    \quad Llama-2-chat & 3.4 & 9.0 / 0.5 \\
    \quad AgentLM & 11.0 & 5.0 / 14.2 \\
    \quad WizardMath & 19.7 & 28.0 / 15.3 \\
    \quad CodeLlama-i & 4.8 & 13.0 / 0.5 \\
    \quad Deepseek-coder-i & 29.0 & 42.0 / 22.1 \\
    \quad Gemini-Pro & 35.2 & 52.0 / 26.3 \\
    \quad GPT-4 & 57.2 & 86.0 / 42.1 \\
    \rowcolor[gray]{0.95} \multicolumn{3}{l}{\textit{Code Interpreter Assistants}}\\
    \quad GPT-3.5 Turbo & 47.2 & 65.0 / 37.9 \\
    \quad GPT-4 & \underline{\textbf{59.7}} & \underline{\textbf{89.0}} / 44.2 \\
    \bottomrule
\end{tabular}
\caption{Performance of models on \textsc{QRText}. Numbers are accuracies in percentages (\%). Best results are in bold, and model-level best results are underlined. The suffix -i indicates -instruct.}
\label{table-results-no-data}
\end{table}

%% file: tables/results_multiple_choice_without_data.tex
\begin{table}[t]
\centering
\small
\begin{tabular}{lcc}
    \toprule
    \textbf{Model} & \textbf{Acc. w/ data} & \textbf{Acc. w/o data} \\
    \midrule
    Random & 35.7 & 35.7 \\
    GPT-3.5 Turbo & \textbf{35.1} & 25.3 \\
    Gemini-Pro & 45.4 & \textbf{48.8} \\
    GPT-4 & 50.7 & \textbf{60.0} \\
    \bottomrule
\end{tabular}
\caption{Performance of models on multiple-choice causal questions of \textsc{QRData} with (w/) and without (w/o) access to the data. Better results are in bold.}
\label{table-multiple-choice-wo-data}
\end{table}

%% file: tables/case.tex
\begin{table*}[ht]
    \centering
    \small
    \scalebox{0.95}{
    \renewcommand{\arraystretch}{1.2}
    \begin{tabularx}{\textwidth}{lX}
    \toprule
    \textbf{Question} & Which cause-and-effect relationship is more likely? Please answer with A, B, or C. \\
    & A. R L5 radiculopathy causes L L1 radiculopathy \quad B. L L1 radiculopathy causes R L5 radiculopathy \\
    & C. No causal relationship exists \\
    \textbf{Gold Answer} & C \\
    \midrule
    \rowcolor[gray]{0.95} \multicolumn{2}{l}{GPT-4 Prediction with Data Access} \\
    \textbf{Step 1:} & \texttt{\# Display the first few records to understand the data structure} \\
    \textbf{Code Interpreter} & \texttt{data.head()} \\
    \textbf{Step 2:} & I will examine the data for the following patterns: \\
    & - If patients with ``R L5 radiculopathy'' also frequently have ``L L1 radiculopathy'' and not the other way around, it may suggest that A is more likely. \\
    & - Conversely, if patients with ``L L1 radiculopathy'' also frequently have ``R L5 radiculopathy'' and not the other way around, it may suggest that B is more likely. \\
    & - If there is no clear pattern or both conditions occur independently, it may suggest that C is more likely.\\
    \textbf{Step 3:} & \texttt{\# Probability of L L1 given R L5} \\
    \textbf{Code Interpreter} & \texttt{P\_L\_L1\_given\_R\_L5 = co\_occurrence\_count / R\_L5\_count if R\_L5\_count else 0} \\
    & \texttt{\# Probability of R L5 given L L1} \\
    & \texttt{P\_R\_L5\_given\_L\_L1 = co\_occurrence\_count / L\_L1\_count if L\_L1\_count else 0} \\
    & \texttt{P\_L\_L1\_given\_R\_L5, P\_R\_L5\_given\_L\_L1} \\
    \textbf{Step 4:} & Based on these conditional probabilities, it is more likely that ``L L1 radiculopathy'' causes `R L5 radiculopathy'' over the other proposed causal direction. Final answer: B. \\
    \rowcolor[gray]{0.95} \multicolumn{2}{l}{GPT-4 Prediction without Data Access} \\
    \textbf{Prediction} & Given both `R L5' and `L L1' radiculopathies are on different sides of the body and different spinal levels, it's not common for one to cause the other according to standard medical understanding. It is more likely that they occur independently due to separate causes. \\
    & It's likely that if both conditions are present in a patient, they arose independently or due to a systemic issue affecting the spine broadly, rather than one causing the other. Final answer: C. \\
    \bottomrule
    \end{tabularx}
    }
    \caption{An example of GPT-4 prediction on \textsc{QRData}. GPT-4 correctly predicts the answer without data access, but predicts wrongly when the data is provided. The outputs are abridged and the full outputs are in Appendix~\ref{sec-appendix-case}.}
    \label{table-case}
\end{table*}

%% file: related_work.tex
\section{Related Work}
\subsection{Data-related Benchmarks} 
\noindent\textbf{Table Question Answering.}
Table QA~\citep{pasupat-liang-2015-compositional, zhong2017seq2sql} asks models to answer questions about information from given tables. It evaluates models' abilities to parse queries, look up information from tables, and integrate the retrieved information to reach an answer. Datasets like Spider~\citep{yu-etal-2018-spider} and SEDE~\citep{hazoom-etal-2021-text} focus more on the semantic parsing part and design complex questions that need compositional generalization over SQL queries. Some datasets present the challenges of retrieving information from hierarchical tables~\citep{cheng-etal-2022-hitab, zhao-etal-2022-multihiertt} or a fusion of textual and tabular data~\citep{chen-etal-2020-hybridqa, zhu-etal-2021-tat}. Datasets like FinQA~\citep{chen-etal-2021-finqa} and \textsc{TabMWP}~\citep{lu2022dynamic} require numerical reasoning to figure out the answer, but they mainly focus on basic arithmetic, while our benchmark requires more advanced quantitative reasoning skills. 

Furthermore, although some benchmarks require multi-hop reasoning, they can be solved in a single turn, whereas our benchmark needs multi-turn reasoning between the model and data.

\vspace{2mm}
\noindent\textbf{Code Generation.}
Code generation tasks require models to produce code to meet certain intents. Recent benchmarks like DS-1000~\citep{lai2023ds} and ExeDS~\citep{huang-etal-2022-execution} explore LLMs' abilities to generate data science code, like training models using machine learning packages and making visualizations. Knowledge\textsc{Math}~\citep{zhao2023knowledgemath} and BizBench~\citep{koncel2023bizbench} evaluate if models could generate code for financial reasoning with tables.
To our knowledge, none of them focus on advanced quantitative reasoning. Only a concurrent dataset DAEval~\citep{hu2024infiagent} involves correlation analysis as a component of question concepts, and none of them include causal reasoning questions.

\subsection{Quantitative Reasoning Abilities}
There is a growing interest in exploring language models' abilities of quantitative reasoning. Various benchmarks focus on mathematical reasoning~\citep{hendrycks2021measuring, cobbe2021training}, and recent works broaden the analysis to the field of science~\citep{lu2022learn, chen-etal-2023-theoremqa}. 

Another line of work focuses on the specific aspect of causal reasoning of LLMs. \citet{bhagavatula2019abductive, qin-etal-2019-counterfactual, liu-etal-2023-magic} explore the ability of commonsense causal reasoning like abductive reasoning and counterfactual reasoning. \citet{willig2022can, gao-etal-2023-chatgpt, kiciman2023causal} analyze the mastery of causal knowledge like cause-effect pairs. \citet{jin2023cladder, jin2023can} evaluate the utilization of causal inference skills.

In this work, we extend the research to quantitative reasoning \emph{with data}, which poses new challenges for LLMs.

%% file: conclusion.tex
\section{Conclusion}
We propose \textsc{QRData}, a benchmark of advanced quantitative reasoning with data, to explore LLMs' statistical and causal reasoning abilities with real-world data. We carefully construct the benchmark with questions from teaching and research materials, and compare natural language reasoning, program-based reasoning, and agent reasoning among a wide range of models. Experimental results reveal that models have difficulties in \emph{data analysis} and \emph{causal reasoning}. Even though powerful LLMs like GPT-4 have acquired causal knowledge, they can hardly integrate them with the provided data. By proposing this benchmark, we hope to draw attention to the data-based quantitative reasoning ability. We call for more specialized methods to enhance the ability, like training models in the agent reasoning style with data as input, and equipping models with more quantitative reasoning knowledge.

%% file: limitation.tex
\section*{Limitations}
\paragraph{Scope of the Benchmark.} We collect questions related to two primary aspects of advanced quantitative reasoning: statistical reasoning and causal reasoning in this benchmark. We focus on the explicit utilization of these reasoning abilities, more specifically, statistical/causal reasoning with data.
And our benchmark is limited to English questions and data. We look forward to future benchmarks on a broader domain and other languages.

\paragraph{Models and Reasoning Methods.} The field of LLM research is developing quickly. We experiment with representative models and reasoning methods in this paper, but we acknowledge that this does not cover all models and frameworks. Methods of constructing training data and finetuning open-source models like Lumos~\citep{yin2023lumos} can also be explored to enhance the effectiveness of models.

\section*{Ethics Statement}
\paragraph{Intellectual Property.} We follow the permission of licenses during the data collection process. And the collected data should not be used commercially.
\paragraph{Potential Risks.} Our evaluation reveals that current LLMs are imperfect in conducting quantitative reasoning with data. We suggest readers not trust solutions provided by LLMs.

%% file: appendix.tex
\appendix
\section{Appendix}
\label{sec-appendix}
\subsection{Question Collection of \textsc{QRData}}
We provide the number of questions collected from different sources in Table~\ref{table-dataset-source}.
Table~\ref{table-topic-example} summarizes the topics covered in \textsc{QRData}, and examples of questions related to these topics.

\label{sec-appendix-qrdata}
\paragraph{Causal Discovery.} We collect both pairwise causal discovery and full graph causal discovery questions from the Neuropathic pain diagnosis dataset following~\citet{kiciman2023causal}. 

We sample 50 causal relations like \emph{L1 radiculopathy $\rightarrow$ IBS} from the gold causal graph, and generate pairwise causal discovery questions asking models to predict the direction of the causal relation. The questions are in the form of \emph{Which cause-and-effect relationship is more likely? A. R L1 radiculopathy causes IBS B. IBS causes R L1 radiculopathy}. 

In full graph causal discovery, there are three possible relations between two variables $A$ and $B$: $A \rightarrow B$, $B \rightarrow A$, and no causal relation exists. We sample 25 variable pairs that have causal relations in the causal graph and 25 variable pairs that do not have causal relations, and form 50 questions in the form of \emph{Which cause-and-effect relationship is more likely? A. headache causes groin pain B. groin pain causes headache C. No causal relationship exists}.

For the Flow cytometry and Arctic sea ice datasets, we only collect the more challenging full graph causal discovery questions. As the causal graph of Arctic sea ice contains double-sided edges, we ask models to predict from four possible choices like \emph{Which cause-and-effect relationship is more likely? A. Residual\_cloud\_cover causes Residual\_longwave B. Residual\_longwave causes Residual\_cloud\_cover C. The causal relation is double-sided between Residual\_cloud\_cover and Residual\_longwave D. No causal relationship exists}.

\paragraph{Causal Estimation.} We collect numerical questions of estimating \emph{average treatment effect (ATE)}, \emph{average treatment effect on the treated (ATT)}, and \emph{average treatment effect on the control (ATC)} from the IHDP dataset. As all the outcomes and counterfactual outcomes are simulated using real covariates, we calculate the values of these estimates as the gold answers.
In the Jobs dataset, all the treatments were from randomized experiments, so we can calculate the true ATT values. We collect questions of estimating ATT from the dataset.

\subsection{Question Collection of \textsc{QRText}}
\label{sec-appendix-qrtext}
\input{tables/dataset_source}
\input{tables/error_analysis}
\input{tables/topic_example}
\begin{figure*}[t]
    \centering
    \includegraphics[width=\linewidth]{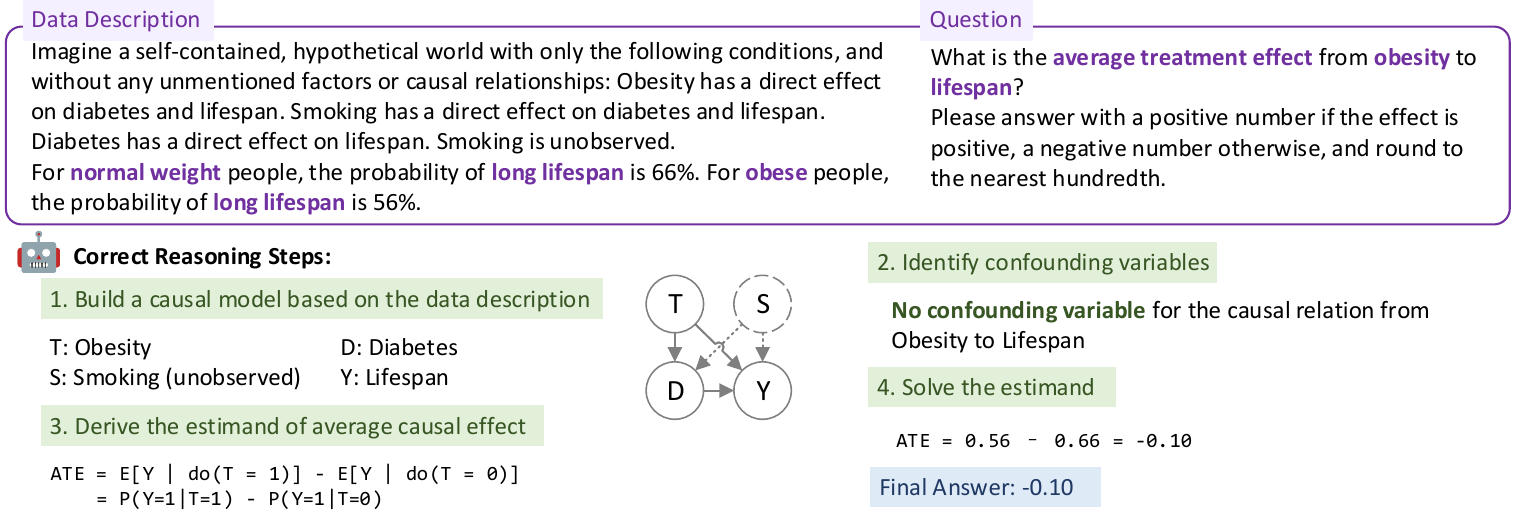}
    \caption{An example question of \textsc{QRText} and correct reasoning steps. The question is converted from \textsc{CLadder}.} 
    \label{fig-qrtext}
\end{figure*}
\textsc{QRText} contains 290 questions in total, 140 multiple choice questions and 150 numerical questions. We collect 100 statistical reasoning questions from OpenIntro Statistics, 94 causal reasoning questions from \textsc{CLadder}, and 96 causal reasoning questions from \textsc{Corr2Cause}. Figure~\ref{fig-qrtext} exhibits an example question of \textsc{QRText}.

The OpenIntro Statistics questions are collected from exercises with solutions from the textbook. \textsc{Corr2Cause} contains causal discovery questions of determining the causal relationship between the variables given correlational statements. We randomly sample \textsc{Corr2Cause} questions from the original test set. 

\textsc{CLadder} contains causal questions where the conditions are described in natural language. It classifies questions into three rungs: associational (rung 1), interventional (rung 2), and counterfactual (rung 3). We only consider questions of rungs 2 and 3 because these are related to causal reasoning. We randomly sample causal estimation questions querying Average Treatment Effect (ATE), Effect of the Treatment on the Treated (ETT), Natural Direct Effect (NDE), and Natural Indirect Effect (NIE), and transform the original yes-or-no questions into numerical questions using the provided metadata. This is to balance the ratio of multiple choice questions and numerical questions, controlling the random performance to be close to \textsc{QRData}.

\subsection{Prompt Design}
\label{sec-appendix-prompt}
\input{tables/prompt_cot}
\input{tables/prompt_answer_extraction}
\input{tables/prompt_pot}
\input{tables/prompt_react}
\input{tables/prompt_assistant}

Table~\ref{table-prompt-cot} shows an example prompt for Chain-of-Thought reasoning. In real prompts, we feed all the content of the data sheets until it meets the length limit of 3500 tokens. We shuffle the data to avoid the bias towards values in the front. After Chain-of-Thought reasoning, we use GPT-3.5 Turbo for answer extraction, and the prompt template is shown in Table~\ref{table-prompt-answer-extraction}.

Table~\ref{table-prompt-pot} shows an example prompt for Program-of-Thoughts reasoning. We shuffle the data and provide the first ten lines in the prompt.
An example prompt for ReAct-style reasoning is in Table~\ref{table-prompt-react}, and an example prompt for code interpreter assistants is in Table~\ref{table-prompt-assistant}. When the question is based on multiple data sheets, we demonstrate the rows of these data sheets respectively in the prompts.

\subsection{Implementation Details}
\label{sec-appendix-implementation}
We conduct model inference with temperature $0$ and maximum new tokens $512$. We limit the maximum reasoning steps to $10$ in ReAct-style prompting.
Experiments are conducted on NVIDIA A800 GPUs. It takes five minutes to evaluate a 7B model on \textsc{QRData} with CoT/PoT prompting on a single GPU, and one hour to evaluate a 7B model with ReAct-style prompting.

\subsection{Case Study}
\label{sec-appendix-case}
\input{tables/case_full_w_data}
\input{tables/case_full_wo_data}
Table~\ref{table-case-full} demonstrates the full output of GPT-4 on the example case from \textsc{QRData}, and Table~\ref{table-case-full-wo} demonstrates the full output of GPT-4 without access to the data.

\subsection{Error Analysis}
\label{sec-appendix-error}
To have a quantitative view of model failures, we conduct an error analysis for the best-performed model GPT-4 with the code interpreter assistant on \textsc{QRData}. We randomly sample 50 incorrect predicted instances, and annotate their main reason for mis-prediction. As shown in Table~\ref{table-error-analysis}, 34\% of the mis-predictions originate from data analysis. The model may fail to retrieve the correct columns according to the instruction of questions, or misinterpret the data values. 44\% of the errors occur in knowledge recall, like choosing a wrong hypothesis test, or using statistical methods to address a causal question. Even if the model recalls the correct knowledge to solve the question, it formulates wrong equations in 16\% of the cases. And 6\% of errors occur in the calculation. This is also the main reason for errors when humans conduct the task, because state-of-the-art causal methods cannot perfectly reach the correct answer.

%% file: tables/dataset_source.tex
\begin{table}
\centering
\small
    \begin{tabular}{lc}
        \toprule
        \textbf{Source} & \textbf{Question} \\
        \midrule
        OpenIntro Statistics & 112  \\
        Quantitative Social Science & 35 \\
        Causal Inference for the Brave and True & 24 \\
        Flow Cytometry & 50 \\
        IHDP & 30 \\
        Jobs & 10 \\
        Neuropathic Pain Diagnosis & 100 \\
        Arctic Sea Ice & 50 \\
        \rowcolor[gray]{0.95} Overall & 411 \\
        \bottomrule
    \end{tabular}
    \caption{Count of questions from different sources of \textsc{QRData}.}
    \label{table-dataset-source}
\end{table}

%% file: tables/error_analysis.tex
\begin{table}[t]
\centering
\small
    \begin{tabular}{lc}
        \toprule
        \textbf{Failure Reason} & \textbf{Percentage (\%)} \\
        \midrule
        Data Analysis & 34\% \\
        Knowledge Recall & 44\% \\
        Equation Formulation & 16\% \\
        Calculation & 6\% \\
        \bottomrule
    \end{tabular}
    \caption{Error analysis of GPT-4 with the code interpreter assistant on \textsc{QRData}.}
    \label{table-error-analysis}
\end{table}

%% file: tables/topic_example.tex
\begin{table*}[t]
\centering
\small
\renewcommand{\arraystretch}{1.2}
\begin{subtable}{\textwidth}
    \centering
    \begin{tabularx}{0.95\textwidth}{>{\hsize=.5\hsize}X>{\hsize=1.25\hsize}X>{\hsize=1.25\hsize}X}
    \toprule
    \textbf{Topic} & \textbf{Description} & \textbf{Example} \\
    \midrule
    Probability & Calculating the likelihood of events happening. & What is the probability a randomly selected person who was not inoculated died from smallpox? \\
    \arrayrulecolor{black!30}\midrule
    Distribution & Understanding how data points are spread out. & Is it reasonable to model the proportion of US adults who support expanding wind turbines using a normal distribution? \\
    \arrayrulecolor{black!30}\midrule
    Estimation & Approximating parameters based on samples drawn from the population, and quantifying the uncertainty of estimates. & Use a Poisson distribution to approximate the data. What is the event rate of the Poisson distribution? \\
    \arrayrulecolor{black!30}\midrule
    Hypothesis Testing & Using statistical tests to assess the validity of hypotheses. & Given the hypothesis that the proportion of college graduates who support off-shore drilling in California is equal to that of non-college graduates, will you accept or reject the hypothesis with a significance level of 5\%? \\
    \arrayrulecolor{black!30}\midrule
    Prediction & Using statistical models to predict future values based on existing data trends. & Fit a linear regression model using the Democratic margin in the two-party vote share as the response variable and the perceived competence for Democratic candidates as the predictor. What is the estimated slope of the model? \\
    \arrayrulecolor{black}\bottomrule
    \end{tabularx}
    \caption{Topics of statistical reasoning.}
\end{subtable}
\begin{subtable}{\textwidth}
    \vspace{1em}%
    \centering
    \begin{tabularx}{0.95\textwidth}{>{\hsize=.5\hsize}X>{\hsize=1.25\hsize}X>{\hsize=1.25\hsize}X}
    \toprule
    \textbf{Topic} & \textbf{Description} & \textbf{Example} \\
    \midrule
    Confounding	& Recognizing confounders that might influence the observed relationship between two variables. For example, a variable that causes both the treatment and the outcome is a confounder.	& We are estimating the causal effect of the new drug on the number of days the patient stays in the hospital, and we already controlled for the severity. Should we also control for the 'hospital' variable? \\
    \arrayrulecolor{black!30}\midrule
    Causal Discovery & Inferring the causal relations from the data. For example, distinguishing between correlation and causation and determining the causal direction between two variables. & Which cause-and-effect relationship is more likely? A. R L5 radiculopathy causes L L1 radiculopathy B. L L1 radiculopathy causes R L5 radiculopathy C. No causal relationship exists \\
    \arrayrulecolor{black!30}\midrule
    Causal Effect Estimation & Estimating the causal effect between two variables using methods like propensity score matching and doubly robust estimation. & What is the average treatment effect of the growth mindset on the achievement score? Please conduct a doubly robust estimation and use logistic regression to estimate the propensity score. \\
    \arrayrulecolor{black!30}\midrule
    Instrumental Variables & Addressing confounding bias with instrumental variables. An instrumental variable causes the treatment and is only correlated with the outcome through the treatment. & What is the average additional percentage wage increase associated with each additional year of education based on the instrumental variable of whether the individual is born in the last quarter (q4)? \\
    \arrayrulecolor{black!30}\midrule
    Panel Data & Applying methods like difference-in-differences and synthetic control to estimate the causal effect in panel data. These methods differ from those used to estimate causal effects in a single period. & What is the average increase in bank deposits per customer in Porto Alegre after the billboard intervention, as estimated by the difference-in-differences approach? \\
    \arrayrulecolor{black}\bottomrule
    \end{tabularx}
    \caption{Topics of causal reasoning.}
\end{subtable}
\caption{Topics covered in \textsc{QRData} and examples of
questions related to these topics.}
\label{table-topic-example}
\end{table*}

%% file: tables/prompt_cot.tex
\begin{table*}[ht]
    \centering
    \small
    \renewcommand{\arraystretch}{1.2}
    \begin{tabularx}{\textwidth}{X}
    \toprule
    Data Description: \\
    Public health has improved and evolved, but has the public's knowledge changed with it? This data set rosling\_responses.csv contains sample responses for two survey questions posed by Hans Rosling during lectures to a wide array of college-educated audiences. \\
    rosling\_responses: \\
     |    | question                            | response   |   prob\_random\_correct | \\
    |  0 | children\_in\_2100                    | correct    |              0.333333 | \\
    |  1 | children\_in\_2100                    | incorrect  |              0.333333 | \\
    |  2 | children\_in\_2100                    | correct    |              0.333333 | \\
    |  3 | children\_with\_1\_or\_more\_vaccination | incorrect  |              0.333333 | \\
    |  4 | children\_in\_2100                    | incorrect  |              0.333333 | \\
    ... \\
    Task: \\
    You are a data analyst and good at quantitative reasoning. You are required to respond to a quantitative question using the provided data. The description and the table are listed above. Please analyze the table to answer the question. Do not write any code in your answer. Ensure that your final answer is positioned at the very end of your output, adhering to the format `Final answer: [answer]'. The final answer should be a number or a short phrase and should be written in a new line. \\
    Question: \\
    Can we reject the notion that the performance of college-educated adults was different than random guessing in the children\_with\_1\_or\_more\_vaccination question with 95\% confidence? Please answer with ``yes'' or ``no''. \\
    Response: \\
    Let's think step by step. \\
    \bottomrule
    \end{tabularx}
    \caption{Example prompt for Chain-of-Thought reasoning. In real prompts, we feed all the content of the data sheets until it meets the length limit.}
    \label{table-prompt-cot}
\end{table*}

%% file: tables/prompt_answer_extraction.tex
\begin{table*}[ht]
    \centering
    \small
    \renewcommand{\arraystretch}{1.2}
    \begin{tabularx}{\textwidth}{X}
    \toprule
    Extract the final answer from the given solution as a numeric value or a short phrase for the question. If you cannot extract an answer, return "None". \\
    You should either return "None" or the final answer without any additional words. \\
    Question: \{question\} \\
    Solution: \{answer\} \\
    Final Answer: \\
    \bottomrule
    \end{tabularx}
    \caption{Prompt template for answer extraction after Chain-of-Thought reasoning. \{question\} is the question from the benchmark, and \{answer\} is the answer generated by models.}
    \label{table-prompt-answer-extraction}
\end{table*}

%% file: tables/prompt_pot.tex
\begin{table*}[ht]
    \centering
    \small
    \renewcommand{\arraystretch}{1.2}
    \begin{tabularx}{\textwidth}{X}
    \toprule
    You are a data analyst and good at quantitative reasoning. You are required to respond to a quantitative question using the provided data. The description and the question can be found below. Please analyze the first 10 rows of the table and write python code to analyze the whole table. You can use any python library. The returned value of the program is supposed to be the answer. The format of the code should be \\
    \`{ }\`{ }\`{ }python \\
    def solution(): \\
    \quad\# import libraries if needed \\
    \quad\# load data \\
    \quad\# write code to get the answer \\
    \quad\# return answer \\
    \`{ }\`{ }\`{ }\\
    Data Description: \\
    Public health has improved and evolved, but has the public's knowledge changed with it? This data set rosling\_responses.csv contains sample responses for two survey questions posed by Hans Rosling during lectures to a wide array of college-educated audiences. \\
    First 10 rows of the data: \\
    rosling\_responses: \\
    |    | question                            | response   |   prob\_random\_correct | \\
    |  0 | children\_in\_2100                    | correct    |              0.333333 | \\
    |  1 | children\_in\_2100                    | incorrect  |              0.333333 | \\
    |  2 | children\_in\_2100                    | correct    |              0.333333 | \\
    |  3 | children\_with\_1\_or\_more\_vaccination | incorrect  |              0.333333 | \\
    |  4 | children\_in\_2100                    | incorrect  |              0.333333 | \\
    |  5 | children\_in\_2100                    | incorrect  |              0.333333 | \\
    |  6 | children\_in\_2100                    | correct    |              0.333333 | \\
    |  7 | children\_in\_2100                    | incorrect  |              0.333333 | \\
    |  8 | children\_in\_2100                    | incorrect  |              0.333333 | \\
    |  9 | children\_in\_2100                    | incorrect  |              0.333333 | \\
    Question: \\
    Can we reject the notion that the performance of college-educated adults was different than random guessing in the children\_with\_1\_or\_more\_vaccination question with 95\% confidence? Please answer with ``yes'' or ``no''. \\
    Response: \\
    \`{ }\`{ }\`{ }python \\
    \bottomrule
    \end{tabularx}
    \caption{Example prompt for Program-of-Thoughts reasoning.}
    \label{table-prompt-pot}
\end{table*}

%% file: tables/prompt_react.tex
\begin{table*}[ht]
    \centering
    \small
    \renewcommand{\arraystretch}{1.2}
    \begin{tabularx}{\textwidth}{X}
    \toprule
    Data Description: \\
    Public health has improved and evolved, but has the public's knowledge changed with it? This data set rosling\_responses.csv contains sample responses for two survey questions posed by Hans Rosling during lectures to a wide array of college-educated audiences. \\
    You are working with a pandas dataframe in Python. The name of the dataframe is `df'. \\
    You should use the tools below to answer the question posed of you: \\
    python\_repl\_ast: A Python shell. Use this to execute python commands. Input should be a valid python command. When using this tool, sometimes output is abbreviated - make sure it does not look abbreviated before using it in your answer. \\
    \\
    Use the following format: \\
    \\
    Question: the input question you must answer \\
    Thought: you should always think about what to do \\
    Action: the action to take, should be python\_repl\_ast \\
    Action Input: the input to the action, should be the code to execute \\
    Observation: the result of the action \\
    ... (this Thought/Action/Action Input/Observation can repeat N times) \\
    Thought: I now know the final answer \\
    Final Answer: the final answer to the original input question, just provide the answer WITHOUT any other words \\
    \\
    Note that you should import the libraries if needed. \\
    DO NOT create any plotting. \\
    For all outputs in code, THE print() function MUST be called. \\
    If you use Action in this step, stop after generating the Action Input and await the execution outcome from python\_repl\_ast. \\
    If you output the Final Answer in this step, do not use Action. \\
    \\
    Here is an example of using the python\_repl\_ast: \\
    Action: python\_repl\_ast \\
    Action Input: \\
    import pandas as pd \\
    print(df.head()) \\
    \\
    This is the result: \\
    | | question | response | prob\_random\_correct | \\
    | 0 | children\_with\_1\_or\_more\_vaccination | correct | 0.333333 | \\
    | 1 | children\_with\_1\_or\_more\_vaccination | correct | 0.333333 | \\
    | 2 | children\_with\_1\_or\_more\_vaccination | incorrect | 0.333333 | \\
    | 3 | children\_with\_1\_or\_more\_vaccination | incorrect | 0.333333 | \\
    | 4 | children\_with\_1\_or\_more\_vaccination | incorrect | 0.333333 | \\
    \\
    Begin! \\
    Question: \\Can we reject the notion that the performance of college-educated adults was different than random guessing in the children\_with\_1\_or\_more\_vaccination question with 95\% confidence? Please answer with ``yes'' or ``no''. \\
    \bottomrule
    \end{tabularx}
    \caption{Example prompt for ReAct-style reasoning.}
    \label{table-prompt-react}
\end{table*}

%% file: tables/prompt_assistant.tex
\begin{table*}[ht]
    \centering
    \small
    \renewcommand{\arraystretch}{1.2}
    \begin{tabularx}{\textwidth}{lX}
    \toprule
    \textbf{Assistant Instruction} & You are a data analyst and good at quantitative reasoning. You are required to respond to a quantitative question using the provided data. The description and the question can be found below. Please just start answering and don't ask me any questions. Put your output in one message. Ensure that your final answer is positioned at the very end of your output, adhering to the format 'Final answer: [answer] \\
    \textbf{User} & Data Description: \\
    & Public health has improved and evolved, but has the public's knowledge changed with it? This data set rosling\_responses.csv contains sample responses for two survey questions posed by Hans Rosling during lectures to a wide array of college-educated audiences. \\
    & Question: \\
    & Can we reject the notion that the performance of college-educated adults was different than random guessing in the children\_with\_1\_or\_more\_vaccination question with 95\% confidence? Please answer with ``yes'' or ``no''. \\
    \bottomrule
    \end{tabularx}
    \caption{Example prompt for the code interpreter assistants.}
    \label{table-prompt-assistant}
\end{table*}

%% file: tables/case_full_w_data.tex
\begin{table*}[ht]
    \centering
    \small
    \renewcommand{\arraystretch}{1.2}
    \begin{tabularx}{\textwidth}{lX}
    \toprule
    \textbf{Data Description} & The neuropathic pain diagnosis mainly consists of symptom diagnosis, pattern diagnosis, and pathophysiological diagnosis. The csv file Neuropathic\_11.csv contains neuropathic pain diagnosis records in the form of tables of which the row represents different patients and the column represents different diagnostic labels. \\
    \textbf{Question} & Which cause-and-effect relationship is more likely? \\
    & A. R L5 radiculopathy causes L L1 radiculopathy \quad B. L L1 radiculopathy causes R L5 radiculopathy \\
    & C. No causal relationship exists \\
    & Please answer with A, B, or C. \\
    \textbf{Gold Answer} & C \\
    \midrule
    \rowcolor[gray]{0.95} \multicolumn{2}{l}{GPT-4 Prediction with Data Access} \\
    \textbf{Step 1:} & \texttt{import pandas as pd} \\
    \textbf{Code Interpreter} & \texttt{\# Reading the dataset} \\
    & \texttt{file\_path = `/mnt/data/Neuropathic\_11.csv'} \\
    & \texttt{data = pd.read\_csv(file\_path)} \\
    & \texttt{\# Display the first few records to understand the data structure} \\
    & \texttt{data.head()} \\
    \textbf{Step 2:} & Let's proceed with analyzing the relevant columns to identify if there are any cause-and-effect relationships between the ``R L5 radiculopathy'' and ``L L1 radiculopathy'' conditions.\\
    & I will start by filtering these two columns from the dataset and then conducting an analysis to identify the likelihood of either of the radiculopathies causing the other. I will examine the data for the following patterns: \\
    & - If patients with ``R L5 radiculopathy'' also frequently have ``L L1 radiculopathy'' and not the other way around, it may suggest that A is more likely. \\
    & - Conversely, if patients with ``L L1 radiculopathy'' also frequently have ``R L5 radiculopathy'' and not the other way around, it may suggest that B is more likely. \\
    & - If there is no clear pattern or both conditions occur independently, it may suggest that C is more likely.\\
    & To do this analysis, I will compute conditional probabilities: \\
    & - P(L L1 | R L5) \\
    & - P(R L5 | L L1) \\
    & Then I will compare these probabilities to make an inference regarding the most likely cause-and-effect relationship. \\
    \textbf{Step 3:} & \texttt{\# Extract the relevant columns} \\
    \textbf{Code Interpreter} & \texttt{R\_L5\_col = `R L5 radiculopathy'} \\
    & \texttt{L\_L1\_col = `L L1 radiculopathy'} \\
    & \texttt{\# Filter the dataset for these two columns} \\
    & \texttt{radiculopathy\_data = data\_corrected[[R\_L5\_col, L\_L1\_col]]} \\
    & \texttt{\# Calculate the occurrence of each condition and their co-occurrence} \\
    & \texttt{R\_L5\_count = radiculopathy\_data[R\_L5\_col].sum()} \\
    & \texttt{L\_L1\_count = radiculopathy\_data[L\_L1\_col].sum()} \\
    & \texttt{co\_occurrence\_count = radiculopathy\_data[(radiculopathy\_data[R\_L5\_col] == 1) \& (radiculopathy\_data[L\_L1\_col] == 1)].shape[0]} \\
    & \texttt{\# Calculate the conditional probabilities} \\
    & \texttt{\# Probability of L L1 given R L5} \\
    & \texttt{P\_L\_L1\_given\_R\_L5 = co\_occurrence\_count / R\_L5\_count if R\_L5\_count else 0} \\
    & \texttt{\# Probability of R L5 given L L1} \\
    & \texttt{P\_R\_L5\_given\_L\_L1 = co\_occurrence\_count / L\_L1\_count if L\_L1\_count else 0} \\
    & \texttt{P\_L\_L1\_given\_R\_L5, P\_R\_L5\_given\_L\_L1} \\
    \textbf{Step 4:} & The conditional probabilities calculated are as follows: \\
    & - P(L L1 | R L5) $\approx$ 0.0075 \\
    & - P(R L5 | L L1) $\approx$ 0.4122 \\
    & These probabilities suggest that the occurrence of ``L L1 radiculopathy'' does not significantly increase when ``R L5 radiculopathy'' is present, given that P(L L1 | R L5) is quite low. On the other hand, the occurrence of ``R L5 radiculopathy'' is notably higher when ``L L1 radiculopathy'' is present, as P(R L5 | L L1) is much higher. \\
    & Therefore, based on these conditional probabilities, it is more likely that ``L L1 radiculopathy'' causes `R L5 radiculopathy'' over the other proposed causal direction. \\
    & Final answer: B \\
    \bottomrule
    \end{tabularx}
    \caption{Full outputs of GPT-4 on the example case from \textsc{QRData}.}
    \label{table-case-full}
\end{table*}

%% file: tables/case_full_wo_data.tex
\begin{table*}[ht]
    \centering
    \small
    \renewcommand{\arraystretch}{1.2}
    \begin{tabularx}{\textwidth}{lX}
    \toprule
    \rowcolor[gray]{0.95} \multicolumn{2}{l}{GPT-4 Prediction without Data Access} \\
    \textbf{Prediction} & To address which cause-and-effect relationship is more likely between `R L5 radiculopathy' and `L L1 radiculopathy', we would generally look for statistical relationships in the data that may imply causality. Specifically, we would examine the frequency with which one condition appears before the other. If `R L5 radiculopathy' generally appears before `L L1 radiculopathy', it would lend support to the idea that A is more likely. Conversely, if `L L1 radiculopathy' appears before `R L5 radiculopathy', we might lean towards B. \\
    & However, without access to the actual data, we can only reason based on general medical knowledge and the understanding of how radiculopathy typically presents. Radiculopathy is a condition resulting from a compressed nerve in the spine that can cause pain, numbness, or weakness in various parts of the body. Typically, this condition is localized and not known to cause another radiculopathy on the opposite side and a different vertebrae level of the body. \\
    & Given both `R L5' and `L L1' radiculopathies are on different sides of the body and different spinal levels, it's not common for one to cause the other according to standard medical understanding. It is more likely that they occur independently due to separate causes. \\
    & However, unilateral radiculopathy doesn't generally cause contralateral radiculopathy at a different spinal level. This is because the nerve root compression that leads to radiculopathy is typically localized, and it would be unusual for such an issue on one side of the spine to directly cause a similar issue on the opposite side at a different level. Therefore, it's likely that if both conditions are present in a patient, they arose independently or due to a systemic issue affecting the spine broadly, rather than one causing the other. \\
    & Therefore, based on a qualitative argument and standard medical understanding, the answer is most likely that there is no causal relationship between the two conditions, which corresponds to option C. \\
    & Final answer: C. \\
    \bottomrule
    \end{tabularx}
    \caption{Full outputs of GPT-4 on the example case from \textsc{QRData} when it does not have access to the data.}
    \label{table-case-full-wo}
\end{table*}

%% file: main.bbl
\begin{thebibliography}{53}
\expandafter\ifx\csname natexlab\endcsname\relax\def\natexlab#1{#1}\fi

\bibitem[{Bai et~al.(2023)Bai, Bai, Chu, Cui, Dang, Deng, Fan, Ge, Han, Huang et~al.}]{bai2023qwen}
Jinze Bai, Shuai Bai, Yunfei Chu, Zeyu Cui, Kai Dang, Xiaodong Deng, Yang Fan, Wenbin Ge, Yu~Han, Fei Huang, et~al. 2023.
\newblock Qwen technical report.
\newblock \emph{arXiv preprint arXiv:2309.16609}.

\bibitem[{Bhagavatula et~al.(2019)Bhagavatula, Le~Bras, Malaviya, Sakaguchi, Holtzman, Rashkin, Downey, Yih, and Choi}]{bhagavatula2019abductive}
Chandra Bhagavatula, Ronan Le~Bras, Chaitanya Malaviya, Keisuke Sakaguchi, Ari Holtzman, Hannah Rashkin, Doug Downey, Wen-tau Yih, and Yejin Choi. 2019.
\newblock Abductive commonsense reasoning.
\newblock In \emph{International Conference on Learning Representations}.

\bibitem[{Bueno~de Mesquita and Fowler(2021)}]{de2021thinking}
Ethan Bueno~de Mesquita and Anthony Fowler. 2021.
\newblock \emph{Thinking clearly with data: A guide to quantitative reasoning and analysis}.
\newblock Princeton University Press.

\bibitem[{Chen et~al.(2022)Chen, Ma, Wang, and Cohen}]{chen2022program}
Wenhu Chen, Xueguang Ma, Xinyi Wang, and William~W Cohen. 2022.
\newblock Program of thoughts prompting: Disentangling computation from reasoning for numerical reasoning tasks.
\newblock \emph{arXiv preprint arXiv:2211.12588}.

\bibitem[{Chen et~al.(2023{\natexlab{a}})Chen, Yin, Ku, Lu, Wan, Ma, Xu, Wang, and Xia}]{chen-etal-2023-theoremqa}
Wenhu Chen, Ming Yin, Max Ku, Pan Lu, Yixin Wan, Xueguang Ma, Jianyu Xu, Xinyi Wang, and Tony Xia. 2023{\natexlab{a}}.
\newblock \href {https://doi.org/10.18653/v1/2023.emnlp-main.489} {{T}heorem{QA}: A theorem-driven question answering dataset}.
\newblock In \emph{Proceedings of the 2023 Conference on Empirical Methods in Natural Language Processing}, pages 7889--7901, Singapore. Association for Computational Linguistics.

\bibitem[{Chen et~al.(2020)Chen, Zha, Chen, Xiong, Wang, and Wang}]{chen-etal-2020-hybridqa}
Wenhu Chen, Hanwen Zha, Zhiyu Chen, Wenhan Xiong, Hong Wang, and William~Yang Wang. 2020.
\newblock \href {https://doi.org/10.18653/v1/2020.findings-emnlp.91} {{H}ybrid{QA}: A dataset of multi-hop question answering over tabular and textual data}.
\newblock In \emph{Findings of the Association for Computational Linguistics: EMNLP 2020}, pages 1026--1036, Online. Association for Computational Linguistics.

\bibitem[{Chen et~al.(2023{\natexlab{b}})Chen, Qian, Tang, Lai, Liu, Han, and Jia}]{chen2023longlora}
Yukang Chen, Shengju Qian, Haotian Tang, Xin Lai, Zhijian Liu, Song Han, and Jiaya Jia. 2023{\natexlab{b}}.
\newblock Longlora: Efficient fine-tuning of long-context large language models.
\newblock \emph{arXiv preprint arXiv:2309.12307}.

\bibitem[{Chen et~al.(2021)Chen, Chen, Smiley, Shah, Borova, Langdon, Moussa, Beane, Huang, Routledge, and Wang}]{chen-etal-2021-finqa}
Zhiyu Chen, Wenhu Chen, Charese Smiley, Sameena Shah, Iana Borova, Dylan Langdon, Reema Moussa, Matt Beane, Ting-Hao Huang, Bryan Routledge, and William~Yang Wang. 2021.
\newblock \href {https://doi.org/10.18653/v1/2021.emnlp-main.300} {{F}in{QA}: A dataset of numerical reasoning over financial data}.
\newblock In \emph{Proceedings of the 2021 Conference on Empirical Methods in Natural Language Processing}, pages 3697--3711, Online and Punta Cana, Dominican Republic. Association for Computational Linguistics.

\bibitem[{Cheng et~al.(2023)Cheng, Li, and Bing}]{cheng2023gpt}
Liying Cheng, Xingxuan Li, and Lidong Bing. 2023.
\newblock Is gpt-4 a good data analyst?
\newblock \emph{arXiv preprint arXiv:2305.15038}.

\bibitem[{Cheng et~al.(2022)Cheng, Dong, Wang, Jia, Guo, Gao, Han, Lou, and Zhang}]{cheng-etal-2022-hitab}
Zhoujun Cheng, Haoyu Dong, Zhiruo Wang, Ran Jia, Jiaqi Guo, Yan Gao, Shi Han, Jian-Guang Lou, and Dongmei Zhang. 2022.
\newblock \href {https://doi.org/10.18653/v1/2022.acl-long.78} {{H}i{T}ab: A hierarchical table dataset for question answering and natural language generation}.
\newblock In \emph{Proceedings of the 60th Annual Meeting of the Association for Computational Linguistics (Volume 1: Long Papers)}, pages 1094--1110, Dublin, Ireland. Association for Computational Linguistics.

\bibitem[{Cobbe et~al.(2021)Cobbe, Kosaraju, Bavarian, Chen, Jun, Kaiser, Plappert, Tworek, Hilton, Nakano et~al.}]{cobbe2021training}
Karl Cobbe, Vineet Kosaraju, Mohammad Bavarian, Mark Chen, Heewoo Jun, Lukasz Kaiser, Matthias Plappert, Jerry Tworek, Jacob Hilton, Reiichiro Nakano, et~al. 2021.
\newblock Training verifiers to solve math word problems.
\newblock \emph{arXiv preprint arXiv:2110.14168}.

\bibitem[{Diez et~al.(2012)Diez, Barr, and Cetinkaya-Rundel}]{diez2012openintro}
David~M Diez, Christopher~D Barr, and Mine Cetinkaya-Rundel. 2012.
\newblock \emph{OpenIntro statistics}.
\newblock OpenIntro Boston, MA, USA:.

\bibitem[{Gao et~al.(2023)Gao, Ding, Qin, and Liu}]{gao-etal-2023-chatgpt}
Jinglong Gao, Xiao Ding, Bing Qin, and Ting Liu. 2023.
\newblock \href {https://doi.org/10.18653/v1/2023.findings-emnlp.743} {Is {C}hat{GPT} a good causal reasoner? a comprehensive evaluation}.
\newblock In \emph{Findings of the Association for Computational Linguistics: EMNLP 2023}, pages 11111--11126, Singapore. Association for Computational Linguistics.

\bibitem[{Guo et~al.(2024)Guo, Zhu, Yang, Xie, Dong, Zhang, Chen, Bi, Wu, Li et~al.}]{guo2024deepseek}
Daya Guo, Qihao Zhu, Dejian Yang, Zhenda Xie, Kai Dong, Wentao Zhang, Guanting Chen, Xiao Bi, Y~Wu, YK~Li, et~al. 2024.
\newblock Deepseek-coder: When the large language model meets programming--the rise of code intelligence.
\newblock \emph{arXiv preprint arXiv:2401.14196}.

\bibitem[{Hazoom et~al.(2021)Hazoom, Malik, and Bogin}]{hazoom-etal-2021-text}
Moshe Hazoom, Vibhor Malik, and Ben Bogin. 2021.
\newblock \href {https://doi.org/10.18653/v1/2021.nlp4prog-1.9} {Text-to-{SQL} in the wild: A naturally-occurring dataset based on stack exchange data}.
\newblock In \emph{Proceedings of the 1st Workshop on Natural Language Processing for Programming (NLP4Prog 2021)}, pages 77--87, Online. Association for Computational Linguistics.

\bibitem[{Hendrycks et~al.(2021)Hendrycks, Burns, Kadavath, Arora, Basart, Tang, Song, and Steinhardt}]{hendrycks2021measuring}
Dan Hendrycks, Collin Burns, Saurav Kadavath, Akul Arora, Steven Basart, Eric Tang, Dawn Song, and Jacob Steinhardt. 2021.
\newblock Measuring mathematical problem solving with the math dataset.
\newblock In \emph{Thirty-fifth Conference on Neural Information Processing Systems Datasets and Benchmarks Track (Round 2)}.

\bibitem[{Hill(2011)}]{hill2011bayesian}
Jennifer~L Hill. 2011.
\newblock Bayesian nonparametric modeling for causal inference.
\newblock \emph{Journal of Computational and Graphical Statistics}, 20(1):217--240.

\bibitem[{Hu et~al.(2024)Hu, Zhao, Wei, Chai, Wang, Wang, Su, Xu, Zhu, Cheng et~al.}]{hu2024infiagent}
Xueyu Hu, Ziyu Zhao, Shuang Wei, Ziwei Chai, Guoyin Wang, Xuwu Wang, Jing Su, Jingjing Xu, Ming Zhu, Yao Cheng, et~al. 2024.
\newblock Infiagent-dabench: Evaluating agents on data analysis tasks.
\newblock \emph{arXiv preprint arXiv:2401.05507}.

\bibitem[{Huang et~al.(2022)Huang, Wang, Zhang, Yan, Cui, Inala, Clement, and Duan}]{huang-etal-2022-execution}
Junjie Huang, Chenglong Wang, Jipeng Zhang, Cong Yan, Haotian Cui, Jeevana~Priya Inala, Colin Clement, and Nan Duan. 2022.
\newblock \href {https://aclanthology.org/2022.dash-1.5} {Execution-based evaluation for data science code generation models}.
\newblock In \emph{Proceedings of the Fourth Workshop on Data Science with Human-in-the-Loop (Language Advances)}, pages 28--36, Abu Dhabi, United Arab Emirates (Hybrid). Association for Computational Linguistics.

\bibitem[{Huang et~al.(2021)Huang, Kleindessner, Munishkin, Varshney, Guo, and Wang}]{huang2021benchmarking}
Yiyi Huang, Matth{\"a}us Kleindessner, Alexey Munishkin, Debvrat Varshney, Pei Guo, and Jianwu Wang. 2021.
\newblock Benchmarking of data-driven causality discovery approaches in the interactions of arctic sea ice and atmosphere.
\newblock \emph{Frontiers in big Data}, 4:642182.

\bibitem[{Imai(2018)}]{imai2018quantitative}
Kosuke Imai. 2018.
\newblock \emph{Quantitative social science: an introduction}.
\newblock Princeton University Press.

\bibitem[{Jin et~al.(2023{\natexlab{a}})Jin, Chen, Leeb, Gresele, Kamal, Zhiheng, Blin, Adauto, Kleiman-Weiner, Sachan et~al.}]{jin2023cladder}
Zhijing Jin, Yuen Chen, Felix Leeb, Luigi Gresele, Ojasv Kamal, LYU Zhiheng, Kevin Blin, Fernando~Gonzalez Adauto, Max Kleiman-Weiner, Mrinmaya Sachan, et~al. 2023{\natexlab{a}}.
\newblock Cladder: A benchmark to assess causal reasoning capabilities of language models.
\newblock In \emph{Thirty-seventh Conference on Neural Information Processing Systems}.

\bibitem[{Jin et~al.(2023{\natexlab{b}})Jin, Liu, Lyu, Poff, Sachan, Mihalcea, Diab, and Sch{\"o}lkopf}]{jin2023can}
Zhijing Jin, Jiarui Liu, Zhiheng Lyu, Spencer Poff, Mrinmaya Sachan, Rada Mihalcea, Mona Diab, and Bernhard Sch{\"o}lkopf. 2023{\natexlab{b}}.
\newblock Can large language models infer causation from correlation?
\newblock \emph{arXiv preprint arXiv:2306.05836}.

\bibitem[{K{\i}c{\i}man et~al.(2023)K{\i}c{\i}man, Ness, Sharma, and Tan}]{kiciman2023causal}
Emre K{\i}c{\i}man, Robert Ness, Amit Sharma, and Chenhao Tan. 2023.
\newblock Causal reasoning and large language models: Opening a new frontier for causality.
\newblock \emph{arXiv preprint arXiv:2305.00050}.

\bibitem[{Koncel-Kedziorski et~al.(2023)Koncel-Kedziorski, Krumdick, Lai, Reddy, Lovering, and Tanner}]{koncel2023bizbench}
Rik Koncel-Kedziorski, Michael Krumdick, Viet Lai, Varshini Reddy, Charles Lovering, and Chris Tanner. 2023.
\newblock Bizbench: A quantitative reasoning benchmark for business and finance.
\newblock \emph{arXiv preprint arXiv:2311.06602}.

\bibitem[{Lai et~al.(2023)Lai, Li, Wang, Zhang, Zhong, Zettlemoyer, Yih, Fried, Wang, and Yu}]{lai2023ds}
Yuhang Lai, Chengxi Li, Yiming Wang, Tianyi Zhang, Ruiqi Zhong, Luke Zettlemoyer, Wen-tau Yih, Daniel Fried, Sida Wang, and Tao Yu. 2023.
\newblock Ds-1000: A natural and reliable benchmark for data science code generation.
\newblock In \emph{International Conference on Machine Learning}, pages 18319--18345. PMLR.

\bibitem[{Liu et~al.(2023)Liu, Yin, Zhang, Feng, and Zhao}]{liu-etal-2023-magic}
Xiao Liu, Da~Yin, Chen Zhang, Yansong Feng, and Dongyan Zhao. 2023.
\newblock \href {https://doi.org/10.18653/v1/2023.findings-acl.574} {The magic of {IF}: Investigating causal reasoning abilities in large language models of code}.
\newblock In \emph{Findings of the Association for Computational Linguistics: ACL 2023}, pages 9009--9022, Toronto, Canada. Association for Computational Linguistics.

\bibitem[{Lu et~al.(2022{\natexlab{a}})Lu, Mishra, Xia, Qiu, Chang, Zhu, Tafjord, Clark, and Kalyan}]{lu2022learn}
Pan Lu, Swaroop Mishra, Tanglin Xia, Liang Qiu, Kai-Wei Chang, Song-Chun Zhu, Oyvind Tafjord, Peter Clark, and Ashwin Kalyan. 2022{\natexlab{a}}.
\newblock Learn to explain: Multimodal reasoning via thought chains for science question answering.
\newblock \emph{Advances in Neural Information Processing Systems}, 35:2507--2521.

\bibitem[{Lu et~al.(2022{\natexlab{b}})Lu, Qiu, Chang, Wu, Zhu, Rajpurohit, Clark, and Kalyan}]{lu2022dynamic}
Pan Lu, Liang Qiu, Kai-Wei Chang, Ying~Nian Wu, Song-Chun Zhu, Tanmay Rajpurohit, Peter Clark, and Ashwin Kalyan. 2022{\natexlab{b}}.
\newblock Dynamic prompt learning via policy gradient for semi-structured mathematical reasoning.
\newblock In \emph{The Eleventh International Conference on Learning Representations}.

\bibitem[{Luo et~al.(2023)Luo, Sun, Xu, Zhao, Lou, Tao, Geng, Lin, Chen, and Zhang}]{luo2023wizardmath}
Haipeng Luo, Qingfeng Sun, Can Xu, Pu~Zhao, Jianguang Lou, Chongyang Tao, Xiubo Geng, Qingwei Lin, Shifeng Chen, and Dongmei Zhang. 2023.
\newblock Wizardmath: Empowering mathematical reasoning for large language models via reinforced evol-instruct.
\newblock \emph{arXiv preprint arXiv:2308.09583}.

\bibitem[{OpenAI(2023)}]{openai2023gpt4}
OpenAI. 2023.
\newblock Gpt-4 technical report.
\newblock \emph{arXiv preprint arXiv:2303.08774}.

\bibitem[{Pasupat and Liang(2015)}]{pasupat-liang-2015-compositional}
Panupong Pasupat and Percy Liang. 2015.
\newblock \href {https://doi.org/10.3115/v1/P15-1142} {Compositional semantic parsing on semi-structured tables}.
\newblock In \emph{Proceedings of the 53rd Annual Meeting of the Association for Computational Linguistics and the 7th International Joint Conference on Natural Language Processing (Volume 1: Long Papers)}, pages 1470--1480, Beijing, China. Association for Computational Linguistics.

\bibitem[{Qin et~al.(2019)Qin, Bosselut, Holtzman, Bhagavatula, Clark, and Choi}]{qin-etal-2019-counterfactual}
Lianhui Qin, Antoine Bosselut, Ari Holtzman, Chandra Bhagavatula, Elizabeth Clark, and Yejin Choi. 2019.
\newblock \href {https://doi.org/10.18653/v1/D19-1509} {Counterfactual story reasoning and generation}.
\newblock In \emph{Proceedings of the 2019 Conference on Empirical Methods in Natural Language Processing and the 9th International Joint Conference on Natural Language Processing (EMNLP-IJCNLP)}, pages 5043--5053, Hong Kong, China. Association for Computational Linguistics.

\bibitem[{Roziere et~al.(2023)Roziere, Gehring, Gloeckle, Sootla, Gat, Tan, Adi, Liu, Remez, Rapin et~al.}]{roziere2023code}
Baptiste Roziere, Jonas Gehring, Fabian Gloeckle, Sten Sootla, Itai Gat, Xiaoqing~Ellen Tan, Yossi Adi, Jingyu Liu, Tal Remez, J{\'e}r{\'e}my Rapin, et~al. 2023.
\newblock Code llama: Open foundation models for code.
\newblock \emph{arXiv preprint arXiv:2308.12950}.

\bibitem[{Sachs et~al.(2005)Sachs, Perez, Pe'er, Lauffenburger, and Nolan}]{sachs2005causal}
Karen Sachs, Omar Perez, Dana Pe'er, Douglas~A Lauffenburger, and Garry~P Nolan. 2005.
\newblock Causal protein-signaling networks derived from multiparameter single-cell data.
\newblock \emph{Science}, 308(5721):523--529.

\bibitem[{Shalit et~al.(2017)Shalit, Johansson, and Sontag}]{shalit2017estimating}
Uri Shalit, Fredrik~D Johansson, and David Sontag. 2017.
\newblock Estimating individual treatment effect: generalization bounds and algorithms.
\newblock In \emph{International conference on machine learning}, pages 3076--3085. PMLR.

\bibitem[{Shi et~al.(2023)Shi, Ajith, Xia, Huang, Liu, Blevins, Chen, and Zettlemoyer}]{shi2023detecting}
Weijia Shi, Anirudh Ajith, Mengzhou Xia, Yangsibo Huang, Daogao Liu, Terra Blevins, Danqi Chen, and Luke Zettlemoyer. 2023.
\newblock Detecting pretraining data from large language models.
\newblock In \emph{NeurIPS 2023 Workshop on Regulatable ML}.

\bibitem[{Team et~al.(2023)Team, Anil, Borgeaud, Wu, Alayrac, Yu, Soricut, Schalkwyk, Dai, Hauth et~al.}]{team2023gemini}
Gemini Team, Rohan Anil, Sebastian Borgeaud, Yonghui Wu, Jean-Baptiste Alayrac, Jiahui Yu, Radu Soricut, Johan Schalkwyk, Andrew~M Dai, Anja Hauth, et~al. 2023.
\newblock Gemini: a family of highly capable multimodal models.
\newblock \emph{arXiv preprint arXiv:2312.11805}.

\bibitem[{Touvron et~al.(2023)Touvron, Martin, Stone, Albert, Almahairi, Babaei, Bashlykov, Batra, Bhargava, Bhosale et~al.}]{touvron2023llama}
Hugo Touvron, Louis Martin, Kevin Stone, Peter Albert, Amjad Almahairi, Yasmine Babaei, Nikolay Bashlykov, Soumya Batra, Prajjwal Bhargava, Shruti Bhosale, et~al. 2023.
\newblock Llama 2: Open foundation and fine-tuned chat models.
\newblock \emph{arXiv preprint arXiv:2307.09288}.

\bibitem[{Tu et~al.(2019)Tu, Zhang, Bertilson, Kjellstrom, and Zhang}]{tu2019neuropathic}
Ruibo Tu, Kun Zhang, Bo~Bertilson, Hedvig Kjellstrom, and Cheng Zhang. 2019.
\newblock Neuropathic pain diagnosis simulator for causal discovery algorithm evaluation.
\newblock \emph{Advances in Neural Information Processing Systems}, 32.

\bibitem[{Vowels et~al.(2022)Vowels, Camgoz, and Bowden}]{vowels2022d}
Matthew~J Vowels, Necati~Cihan Camgoz, and Richard Bowden. 2022.
\newblock D’ya like dags? a survey on structure learning and causal discovery.
\newblock \emph{ACM Computing Surveys}, 55(4):1--36.

\bibitem[{Wei et~al.(2022)Wei, Wang, Schuurmans, Bosma, Xia, Chi, Le, Zhou et~al.}]{wei2022chain}
Jason Wei, Xuezhi Wang, Dale Schuurmans, Maarten Bosma, Fei Xia, Ed~Chi, Quoc~V Le, Denny Zhou, et~al. 2022.
\newblock Chain-of-thought prompting elicits reasoning in large language models.
\newblock \emph{Advances in Neural Information Processing Systems}, 35:24824--24837.

\bibitem[{Willig et~al.(2022)Willig, Ze{\v{c}}evi{\'c}, Dhami, and Kersting}]{willig2022can}
Moritz Willig, Matej Ze{\v{c}}evi{\'c}, Devendra~Singh Dhami, and Kristian Kersting. 2022.
\newblock Can foundation models talk causality?
\newblock In \emph{UAI 2022 Workshop on Causal Representation Learning}.

\bibitem[{Yao et~al.(2022)Yao, Zhao, Yu, Du, Shafran, Narasimhan, and Cao}]{yao2022react}
Shunyu Yao, Jeffrey Zhao, Dian Yu, Nan Du, Izhak Shafran, Karthik~R Narasimhan, and Yuan Cao. 2022.
\newblock React: Synergizing reasoning and acting in language models.
\newblock In \emph{The Eleventh International Conference on Learning Representations}.

\bibitem[{Yin et~al.(2023)Yin, Brahman, Ravichander, Chandu, Chang, Choi, and Lin}]{yin2023lumos}
Da~Yin, Faeze Brahman, Abhilasha Ravichander, Khyathi Chandu, Kai-Wei Chang, Yejin Choi, and Bill~Yuchen Lin. 2023.
\newblock Lumos: Learning agents with unified data, modular design, and open-source llms.
\newblock \emph{arXiv preprint arXiv:2311.05657}.

\bibitem[{Yu et~al.(2018)Yu, Zhang, Yang, Yasunaga, Wang, Li, Ma, Li, Yao, Roman, Zhang, and Radev}]{yu-etal-2018-spider}
Tao Yu, Rui Zhang, Kai Yang, Michihiro Yasunaga, Dongxu Wang, Zifan Li, James Ma, Irene Li, Qingning Yao, Shanelle Roman, Zilin Zhang, and Dragomir Radev. 2018.
\newblock \href {https://doi.org/10.18653/v1/D18-1425} {{S}pider: A large-scale human-labeled dataset for complex and cross-domain semantic parsing and text-to-{SQL} task}.
\newblock In \emph{Proceedings of the 2018 Conference on Empirical Methods in Natural Language Processing}, pages 3911--3921, Brussels, Belgium. Association for Computational Linguistics.

\bibitem[{Yue et~al.(2023)Yue, Qu, Zhang, Fu, Huang, Sun, Su, and Chen}]{yue2023mammoth}
Xiang Yue, Xingwei Qu, Ge~Zhang, Yao Fu, Wenhao Huang, Huan Sun, Yu~Su, and Wenhu Chen. 2023.
\newblock Mammoth: Building math generalist models through hybrid instruction tuning.
\newblock \emph{arXiv preprint arXiv:2309.05653}.

\bibitem[{Zeng et~al.(2023)Zeng, Liu, Lu, Wang, Liu, Dong, and Tang}]{zeng2023agenttuning}
Aohan Zeng, Mingdao Liu, Rui Lu, Bowen Wang, Xiao Liu, Yuxiao Dong, and Jie Tang. 2023.
\newblock Agenttuning: Enabling generalized agent abilities for llms.
\newblock \emph{arXiv preprint arXiv:2310.12823}.

\bibitem[{Zhang et~al.(2023)Zhang, Yue, Li, and Sun}]{zhang2023tablellama}
Tianshu Zhang, Xiang Yue, Yifei Li, and Huan Sun. 2023.
\newblock Tablellama: Towards open large generalist models for tables.
\newblock \emph{arXiv preprint arXiv:2311.09206}.

\bibitem[{Zhao et~al.(2022)Zhao, Li, Li, and Zhang}]{zhao-etal-2022-multihiertt}
Yilun Zhao, Yunxiang Li, Chenying Li, and Rui Zhang. 2022.
\newblock \href {https://doi.org/10.18653/v1/2022.acl-long.454} {{M}ulti{H}iertt: Numerical reasoning over multi hierarchical tabular and textual data}.
\newblock In \emph{Proceedings of the 60th Annual Meeting of the Association for Computational Linguistics (Volume 1: Long Papers)}, pages 6588--6600, Dublin, Ireland. Association for Computational Linguistics.

\bibitem[{Zhao et~al.(2023)Zhao, Liu, Long, Zhang, Zhao, and Cohan}]{zhao2023knowledgemath}
Yilun Zhao, Hongjun Liu, Yitao Long, Rui Zhang, Chen Zhao, and Arman Cohan. 2023.
\newblock Knowledgemath: Knowledge-intensive math word problem solving in finance domains.
\newblock \emph{arXiv preprint arXiv:2311.09797}.

\bibitem[{Zhong et~al.(2017)Zhong, Xiong, and Socher}]{zhong2017seq2sql}
Victor Zhong, Caiming Xiong, and Richard Socher. 2017.
\newblock Seq2sql: Generating structured queries from natural language using reinforcement learning.
\newblock \emph{arXiv preprint arXiv:1709.00103}.

\bibitem[{Zhu et~al.(2021)Zhu, Lei, Huang, Wang, Zhang, Lv, Feng, and Chua}]{zhu-etal-2021-tat}
Fengbin Zhu, Wenqiang Lei, Youcheng Huang, Chao Wang, Shuo Zhang, Jiancheng Lv, Fuli Feng, and Tat-Seng Chua. 2021.
\newblock \href {https://doi.org/10.18653/v1/2021.acl-long.254} {{TAT}-{QA}: A question answering benchmark on a hybrid of tabular and textual content in finance}.
\newblock In \emph{Proceedings of the 59th Annual Meeting of the Association for Computational Linguistics and the 11th International Joint Conference on Natural Language Processing (Volume 1: Long Papers)}, pages 3277--3287, Online. Association for Computational Linguistics.

\end{thebibliography}
